\documentclass[sn-nature]{sn-jnl} 

\usepackage{graphicx}%
\usepackage{multirow}%
\usepackage{amsmath,amssymb,amsfonts}%
\usepackage{amsthm}%
\usepackage{mathrsfs}%
\usepackage[title]{appendix}%
\usepackage{xcolor}%
\usepackage{textcomp}%
\usepackage{manyfoot}%
\usepackage{booktabs}%
\usepackage{algorithm}%
\usepackage{algorithmicx}%
\usepackage{algpseudocode}%
\usepackage{listings}%
\usepackage{appendix}
\usepackage{tcolorbox}

\theoremstyle{thmstyleone}%
\theoremstyle{thmstyletwo}%

\theoremstyle{thmstylethree}%

\begin{document}

\title[Article Title]{OpenSTARLab: Open Approach for Spatio-Temporal Agent Data Analysis in Soccer}


\author[1]{\fnm{Calvin} \sur{Yeung}}\email{yeung.chikwong@g.sp.m.is.nagoya-u.ac.jp}
\author[1]{\fnm{Kenjiro} \sur{Ide}}\email{ide.kenjiro@g.sp.m.is.nagoya-u.ac.jp}
\author[2]{\fnm{Taiga} \sur{Someya}}\email{taiga98-0809@g.ecc.u-tokyo.ac.jp}
\author*[1,3]{\fnm{Keisuke} \sur{Fujii}}\email{fujii@i.nagoya-u.ac.jp}

\affil[1]{\orgdiv{Graduate School of Informatics}, \orgname{Nagoya University}, \orgaddress{\city{Nagoya}, \country{Japan}}}

\affil[2]{\orgdiv{Graduate School of Arts and Sciences}, \orgname{The University of Tokyo}, \orgaddress{\city{Tokyo}, \country{Japan}}}

\affil[3]{\orgdiv{Center for Advanced Intelligence Project}, \orgname{RIKEN}, \orgaddress{\city{Osaka}, \country{Japan}}}


\abstract{Sports analytics has become both more professional and sophisticated, driven by the growing availability of detailed performance data. This progress enables applications such as match outcome prediction, player scouting, and tactical analysis. In soccer, the effective utilization of event and tracking data is fundamental for capturing and analyzing the dynamics of the game. However, there are two primary challenges: the limited availability of event data primarily restricted to top-tier teams and leagues, and the scarcity and high cost of tracking data, which complicates its integration with event data for comprehensive analysis. Here we propose OpenSTARLab, an open-source framework designed to democratize spatio-temporal agent data analysis in sports by addressing these key challenges. OpenSTARLab includes the Pre-processing Package that standardizes event and tracking data through Unified and Integrated Event Data and State-Action-Reward formats, the Event Modeling Package that implements deep learning-based event prediction, alongside the RLearn Package for reinforcement learning tasks. These technical components facilitate the handling of diverse data sources and support advanced analytical tasks, thereby enhancing the overall functionality and usability of the framework. To assess OpenSTARLab’s effectiveness, we conducted several experimental evaluations. These demonstrate the superior performance of the specific event prediction model in terms of action and time prediction accuracies and maintained its robust event simulation performance. Furthermore, reinforcement learning experiments reveal a trade-off between action accuracy and temporal difference loss and show comprehensive visualization. Overall, OpenSTARLab serves as a robust platform for researchers and practitioners, enhancing innovation and collaboration in the field of soccer data analytics.}




\maketitle

\section{Introduction}
\label{sec:introduction}
The increasing availability of sports data has developed analytics, driving applications such as match result prediction \citep{berrar2024data, yeung2024evaluating,yeung2023framework}, player scouting \citep{decroos2019actions,spearman2018beyond,liu2020deep}, and tactical analysis \citep{simpson2022seq2event,baron2024foundation,yeung2023events} through data-driven approaches. A cornerstone of these advancements in soccer analytics is the utilization of match data, particularly the event data and tracking data, which are essential for capturing and analyzing the game's dynamics. Event data records key occurrences such as passes, shots, and fouls and tracking data captures player and ball positions over time. The growing availability of such data, combined with advancements in machine learning techniques, has led to the development of numerous methods for play evaluation using predictive modeling (e.g., \cite{decroos2019actions,simpson2022seq2event}) and reinforcement learning (RL; e.g., \cite{Liu2020,nakahara2023action}). Here we call the data for such analytical tools \emph{spatio-temporal agent data}, and we focus on soccer, a sport characterized by significant investment in data collection and the presence of several prominent data providers.

Currently, two primary challenges hinder the universal accessibility of spatio-temporal agent data analysis.
First, event data, which forms the foundation of many soccer analytics tasks, is sometimes publicly available from data provider companies, but predominantly available only for top-tier teams and leagues. This limitation creates a significant barrier for researchers and analysts working on amateur competitions, where such data is scarce. 
It should be noted that video footage is primarily effective for understanding the game. However, its accessibility is often limited by copyright and privacy concerns, which restrict its widespread use in research. While open-source datasets like SoccerNet \cite{giancola2018soccernet, deliege2021soccernet, cioppa2024soccernet} provide a substantial amount of publicly available footage, though sufficient for computer vision tasks such as tracking \cite{cui2023sportsmot,scott2022soccertrack,scott2024teamtrack} and event detection \cite{giancola2018soccernet,deliege2021soccernet, cioppa2024soccernet,wu2024sportshhi,yeung2024autosoccerpose}, they often lack the extensive annotations required for predictive modeling and RL applications.
Regardless of whether the data is richly annotated or limited in quantity, there remains a lack of accessible infrastructure for obtaining data on amateur teams, often the focus of interest for many researchers and practitioners. To democratize sports analytics, it is essential to establish an open-source platform from annotating data of video footage to training machine learning models.

Second, tracking data is even more limited than event data, with publicly available datasets being exceptionally scarce, which remains primarily accessible only to professional teams through data providers due to higher measurement costs. 
Due to the limitation of access, integrating tracking data with event data presents significant challenges. Although temporal synchronization methods have been discussed in some approaches (e.g., \cite{van2023etsy}), there is a notable lack of accessible pipelines for preprocessing tracking and event data, especially for tasks that require feeding data at each time frame into deep learning models. This limitation is particularly pronounced in tasks involving deep RL (DRL), where data from every time frame is crucial for modeling \cite{nakahara2023action} (for details, see Section \ref{sec:related_work}). 

\begin{figure}
\centering
\includegraphics[width=\textwidth]{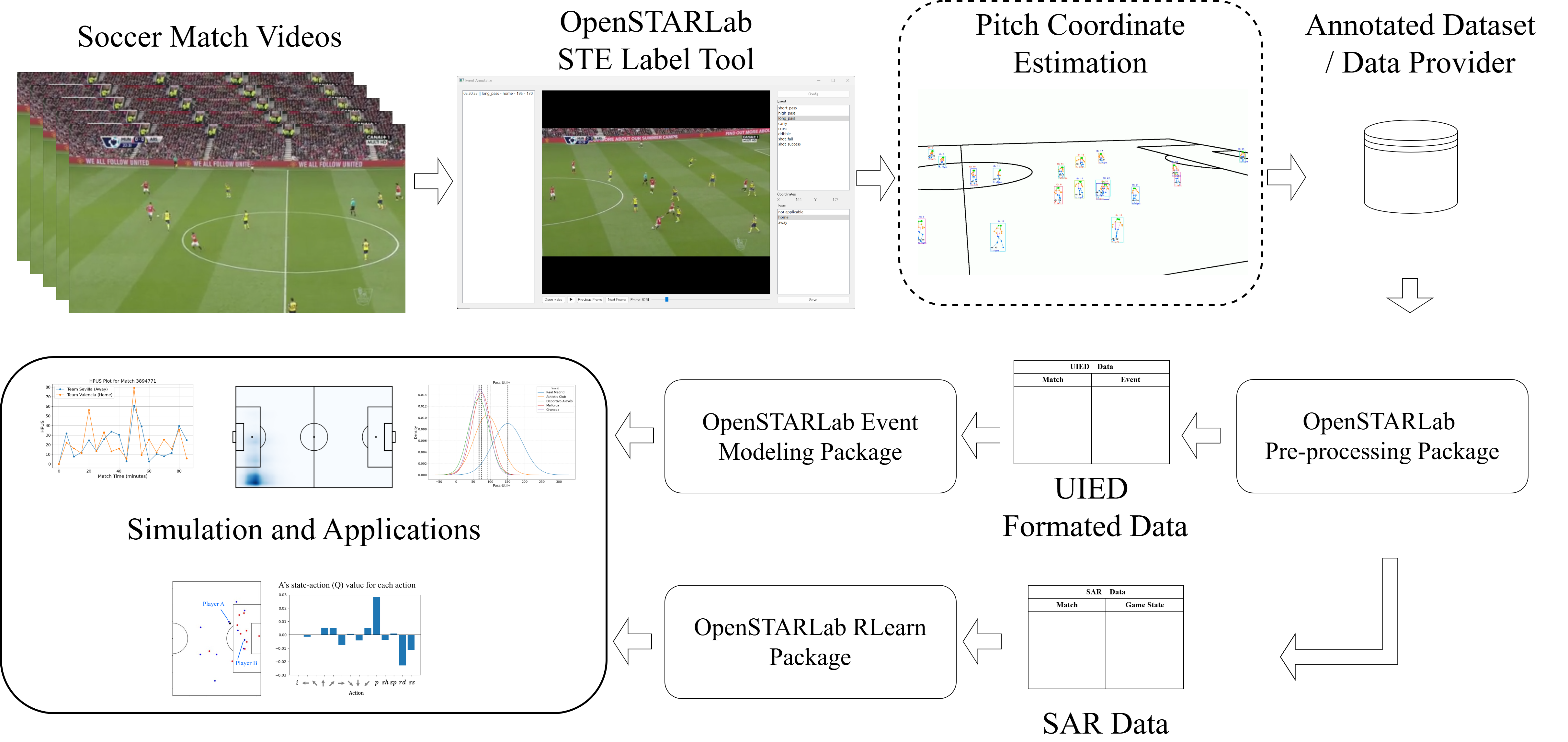}
\caption{Overview of the OpenSTARLab application on soccer. The dashed rectangle indicates steps that may be required depending on the specific soccer match video. Soccer match image from SoccerNet \citep{deliege2021soccernet}.}
\label{fig:openstarlab}
\end{figure}

To address these issues, we propose OpenSTARLab (i.e., \textit{Open approach Spatio-Temporal Agent data Research platform}), a unified framework designed to streamline event annotation, data standardization, and various deep learning modeling for soccer analytics as illustrated in Figure \ref{fig:openstarlab}.
For the first challenge, OpenSTARLab provides an \textit{Event Modeling Package}\footnote{\label{fn:event}OpenSTARLab Event Package: \url{https://github.com/open-starlab/Event}}, which offers a holistic workflow for deep learning-based event prediction models. This package provides tools for model training, applications, and visualization, enabling effective interpretation and analysis of results. To accommodate the diverse data formats from different providers, the platform introduces a \textit{Unified and Integrated Event Data} (UIED) format in \textit{Preprocessing Package}\footnote{\label{fn:preprocessing}OpenSTARLab Pre-processing Package: \url{https://github.com/open-starlab/PreProcessing}} to standardize and prepare the data. These tools ensure that data from both public and private sources can be seamlessly integrated. For cases where annotated data is unavailable, OpenSTARLab also offers an \textit{Spatial-Temporal Event (STE) label Tool}\footnote{\label{fn:event_label_tool}OpenSTARLab STE label Tool: \url{https://github.com/open-starlab/Event_Data_LabelTool}}, enabling users to label video data with event details efficiently, thereby democratizing the ability to prepare data for machine learning tasks.

For the second challenge, OpenSTARLab extends its Preprocessing Package to handle tracking data and introduces the SAR \textit{(State-Action-Reward)} format specifically tailored for deep learning tasks, including DRL. The SAR format enables the sequential processing of data at each time frame, which is critical for modeling tasks that involve temporal dynamics. To further support DRL, the platform includes the \textit{RLearn} Package\footnote{\label{fn:rlearn}OpenSTARLab RLearn Package: \url{https://github.com/open-starlab/RLearn}}, which provides an integrated pipeline for training DRL models, from data preparation to visualization of results. By aligning event and tracking data within a unified framework, OpenSTARLab enables seamless integration and supports advanced DRL applications that leverage spatiotemporal information.

The contributions of this study are summarized as follows. OpenSTARLab offers a comprehensive and unified framework that democratizes spatio-temporal agent data analysis by effectively addressing key challenges in data accessibility, preprocessing, and modeling. This enables researchers and practitioners to conduct advanced analytics and provides greater innovation and collaboration within the field of soccer data science.
The framework includes specialized packages such as the Preprocessing Package, which standardizes event and tracking data through the UIED and SAR formats, and the STE Label Tool, which facilitates efficient event annotation from video data. These technical components provide solutions for handling diverse data sources and support sophisticated deep learning and reinforcement learning tasks.
Experimental evaluations demonstrate the superior performance of the specific event prediction model in terms of action and time prediction accuracies and maintained its robust event simulation performance. Reinforcement learning experiments reveal a trade-off between action accuracy and temporal difference loss and show comprehensive visualization.

The rest of this paper is organized as follows: Section \ref{sec:related_work} reviews related work; Section \ref{sec:openstarlab} introduces the OpenSTARLab framework; Section \ref{sec:experiment} presents experimental results; Section \ref{sec:conclusion} concludes this paper with future directions.

\section{Related work}
\label{sec:related_work}
In this section, the literature on event annotation, data formatting, and predictive modeling for soccer is reviewed, with attention to recent advancements and existing challenges. Section \ref{ssec:event_annotation_in_soccer} examines methods for soccer event annotation, including semi-automatic, automatic, and deep-learning approaches. Section \ref{ssec:event_data_format_in_soccer} addresses the diversity of soccer event data formats across providers. Finally, Sections \ref{ssec:predictive_modeling_for_soccer_event} and \ref{ssec:related_rl_soccer} review predictive modeling and RL approaches in soccer.

\subsection{Event annotation in soccer}
\label{ssec:event_annotation_in_soccer}

In soccer, analyzing player interactions and game events is essential for strategic insights. However, the sport’s complexity and fast pace make accurate event annotation challenging. Reliable and detailed event data, including event types and spatiotemporal information, is crucial for coaches, analysts, and players. Over the years, various methods have been developed to automate the detection and annotation of soccer events from broadcast videos.

Initial approaches relied on semi-automatic methods, which include clustering techniques \citep{ballan2010semantic} and rule-based algorithms \citep{stein2016director}, often enhanced with external text sources \citep{alan2008ontological}. These approaches required extensive domain knowledge, as rule accuracy and cluster reliability are sensitive to factors like camera angles and filming conditions. Subsequently, automatic methods were introduced to improve efficiency. These leveraged audiovisual features and fuzzy rule-based reasoning systems \citep{hosseini2013fuzzy}, visual cues paired with finite state machines \citep{assfalg2003semantic}, and synchronized match reports and video analysis \citep{wang2016soccer}.

The rapid advancement of deep learning has further expanded the capabilities of automatic event detection in soccer. Recent approaches include CNN-based methods \citep{theagarajan2020automated,cioppa2020context}, combinations of inertial sensor data with Random Forest classifiers \citep{barra2023footapp}, self-attention mechanisms \citep{ma2020event}, and hybrid CNN-RNN architectures \citep{sorano2021automatic}. Since 2021, the SoccerNet Challenge \citep{cioppa2024soccernet2024challengesresults} has provided a competitive platform for testing and improving these techniques. Additionally, simulated data has been explored to enhance event detection \citep{morra2020soccer}.

Despite these advancements, both semi-automatic and automatic methods face limitations. Semi-automatic methods require deep domain expertise in both computer science and soccer to ensure the accuracy of rules and clusters, and their performance can be highly affected by variations in video capturing. On the other hand, automatic methods struggle with precision, with the current state-of-the-art method in SoccerNet Challenge 2024 \citep{cioppa2024soccernet2024challengesresults} achieving only a mean average precision of 73. A significant gap in most approaches is the lack of attention to spatial elements in event data, even though modern datasets often include both event types and spatiotemporal information.

Currently, no publicly available annotation tool allows users to comprehensively label event types along with their spatial and temporal dimensions. To address this, we propose the STE Label Tool\footref{fn:event_label_tool}, designed to support the detailed and accurate annotation of event types and integrate spatiotemporal data for more holistic analysis. This tool aims to bridge the current gap in soccer event data annotation, providing an essential resource for researchers and analysts seeking accurate and spatially aware event data for all types of soccer matches.

\subsection{Event data format in soccer}
\label{ssec:event_data_format_in_soccer}

In professional soccer, numerous sports data vendors provide human-annotated event data, which is indispensable for modern soccer analytics, spanning applications in performance analysis, fan engagement, and machine learning-based research. However, using event data from multiple providers presents several challenges \citep{decroos2020vaep}:

\begin{enumerate}
\item \textbf{Diverse Objectives}: Vendors structure data differently to meet diverse objectives, such as audience engagement, in-depth performance analytics, and research, leading to variability in data formats.

\item \textbf{Inconsistent Terminology}: Each vendor uses distinct terminologies and definitions to describe in-game events, making it challenging to standardize the data for comparative or integrative studies.

\item \textbf{Fragmented Event Information}: Many vendors offer additional, optional information for specific event types, creating variability in the data and complicating the application of automated analysis tools consistently across events.

\end{enumerate}

Existing Python packages SoccerAction (SPDAL) \citep{decroos2020vaep} and Kloppy\footnote{Kloppy: \url{https://github.com/PySport/kloppy/tree/master}} have aimed to address these issues by grouping event data into broader categories, facilitating standardized event handling. However, these packages still face limitations: common actions, such as different pass types, are often aggregated under a single category, while rare actions (e.g., penalty shots) are classified separately, limiting flexibility in data analysis \citep{yeung2024unveiling}. Moreover, deep learning and statistical analyses benefit from features that capture transitions between events, which are underdeveloped in these tools.

To address these limitations, we introduce the Unified and Integrated Event Data (UIED) format, implemented in the OpenSTARLab Pre-processing package\footref{fn:preprocessing}. UIED unifies data from various vendors—such as StatsBomb, Wyscout, and DataStadium—into a consistent format that also aligns with the Google Research Football (GRF) RL environment \citep{kurach2020google}. The UIED format integrates essential spatial and temporal features, such as changes in distance, time, and the distance and angle to goal, which have proven valuable in previous studies \citep{yeung2024unveiling,yeung2023events,yeung2023transformer,simpson2022seq2event}. These features support advanced analyses without dependence on optional, vendor-specific fields, making UIED a robust tool for consistent event data analysis. Further details on the UIED format are provided in Section \ref{ssec:UIED_Format}.

\subsection{Predictive modeling for soccer event}
\label{ssec:predictive_modeling_for_soccer_event}

\begin{table}[h]
\centering
\begin{tabular}{lllll}
\toprule
\parbox{2cm}{ \textbf{Model} \\ \textbf{(Year)}}    & \textbf{Data}       & \textbf{Input Variables}                 & \parbox{3cm}{ \textbf{Model Architecture} \\ \textbf{(Cost Function)}}           & \parbox{3cm}{ \textbf{Target Variables} \\ \textbf{(Event at $t+1$)}}            \\ 
\midrule
Seq2Event \citep{simpson2022seq2event}        & Wyscout \cite{pappalardo2019public}            & - xy coordinates                & Transformer encoder,            & - xy coordinates             \\ 
(2022)                 &                     & - action                        & MLP decoder                     & - action          \\ 
                 &                     & - time                          & (CE + RMSE)                              &                              \\ 
                 &                     & - score advantage               &                                 &                              \\ 
                 &                     & - derived features\footnotemark[2]              &                                 &                              \\ 
\hline
NMSTPP \citep{yeung2023transformer}           & Wyscout \cite{pappalardo2019public}            & - zone                          & Transformer encoder,            & - zone                      \\ 
(2023)                 &                     & - action                        & NPP decoder                     & - action                    \\ 
                 &                     & - inter-event time              & (CE + RMSE)                               & - inter-event time \\ 
                 &                     & - derived features\footnotemark[2]              &                                 &                              \\ 
\hline
NMSTPP+360      & StatsBomb           & - zone                          & Transformer encoder,            & - zone                      \\ 
\citep{yeung2024unveiling,yeung2023events}                  &   360 Data                  & - action                        & NPP decoder                     & - action                    \\ (2023, 2024)
                 &                     & - inter-event time              & (CE + RMSE)                                & - inter-event time \\ 
                 &                     & - player coordinates            &                                 &                              \\ 
                 &                     & - derived features\footnotemark[2]              &                                 &                              \\ 
\hline
LEM \citep{mendes2024towards,mendes2024forecasting}            & Wyscout \cite{pappalardo2019public}            & - xy coordinates                & 3 MLPs (non end-to-end)                          & - xy coordinates            \\ 
(2024, 2024)                 &                     & - action                        & (CE)                              & - action                    \\ 
                 &                     & - time                          &                                 & - time                      \\ 
                 &                     & - home team indicator                         &                                 & - home team indicator                   \\ 
                 &                     & - accuracy indicator                      &                                 & - accuracy indicator    \\ 
                 &                     & - goal indicator                &                                 & - goal indicator              \\ 
                 &                     & - match scores                        &                                 &                              \\ 
\hline
FMS \citep{baron2024foundation}              & StatsBomb            & - xy coordinates                & Transformer decoder             & - action        \\ 
 (2024)                & Open Data\footnotemark[1]                    & - action                        & (CE)                              &                              \\ 
                 &                     & - team                          &                                 &                              \\ 
\botrule
\end{tabular}
\caption{Comparison of soccer event prediction models in recent studies. The models are ranked by publication year, and studies that use similar models but differ only in factors other than architecture, inputs, target, or cost function are grouped into the same row. In the model architecture and cost function, MLP denotes multi-layer perceptron, NPP denotes neural point process, CE denotes cross-entropy, and RMSE denotes root mean square error.}
\footnotetext[1]{StatsBomb open data available at \url{https://github.com/statsbomb/open-data}}
\footnotetext[2]{Derived features include the change in time and distance, angle to goal, and distance to goal.}
\label{Tab:compare_event_model}
\end{table}

Predictive modeling in soccer provides quantitative insights into tactical dynamics, team performance, and potential outcomes by forecasting on-field events. A prominent example is the expected goals (xG) model \citep{eggels2016expected}, which predicts the probability of a goal. This concept has been extended to models such as expected assists (xA)\footnote{Expected Assist Model \url{https://www.statsperform.com/opta-analytics/}} and expected threat (xT)\footnote{Expected Threat Model \url{https://karun.in/blog/expected-threat.html}}, which evaluate the likelihood of creating scoring opportunities or increasing the potential threat of a play.

The Valuing Actions by Estimating Probabilities (VAEP) framework \citep{decroos2019actions} further enhances predictive capabilities by modeling the likelihood of scoring or conceding in future time frames. Variants such as VDEP \citep{toda2022evaluation} and GVDEP \citep{umemoto2022location} extend these principles to focus specifically on defensive evaluations. However, these traditional approaches often lack comprehensive representations of future events, limiting their predictive scope. Recent advances in deep learning, particularly with transformer models \citep{vaswani2017attention}, have revolutionized soccer analytics by enabling the capture of complex sequences in match data. These methods provide more robust and comprehensive predictions of future events, offering new possibilities for analyzing and optimizing team performance.

The Seq2Event model \citep{simpson2022seq2event} marked a foundational step in this field. Utilizing a transformer encoder for feature embedding and a multi-layer perceptron (MLP) for decoding, Seq2Event predicts the xy coordinates and action probabilities at the next time step t+1. This model introduced a novel metric, the possession utility (poss-util) score, which assesses team possession.

Building on Seq2Event, the Neural Marked Spatio-Temporal Point Process (NMSTPP) model \citep{yeung2023transformer} integrates a neural marked spatio-temporal point process \citep{Shchur2021} framework. Replacing the MLP decoder, NMSTPP predicts event timing and spatial attributes sequentially, extending the possession utility score into a Holistic Possession Utilization Score (HPUS) by adding temporal considerations. The NMSTPP+360 \citep{yeung2023events} further enhances predictive accuracy by incorporating player coordinate data, and its potential in RL applications is explored in \citep{yeung2024unveiling}.

Further advancements include the  Large Events Model (LEM) model \citep{mendes2024towards}, which forecasts soccer events by independently training three MLPs to predict event attributes sequentially. LEM also introduces the Expected Points Added (xP+) metric to quantify a player’s contribution toward match outcomes. The model has demonstrated potential for both short- and long-term match simulations \citep{mendes2024forecasting}. Lastly, the Foundation Model for Soccer (FMS) \citep{baron2024foundation}, a transformer decoder-based model, predicts upcoming actions with potential applications in action simulations. Table \ref{Tab:compare_event_model} summarizes these soccer event prediction models.

Despite these advancements, existing studies use varying frameworks and evaluation metrics, complicating direct comparisons. Additionally, most models lack comprehensive simulation performance evaluations. To address these gaps, we introduce the UIED format alongside the OpenStarLab Event package\footref{fn:event}. Together, they establish a standardized platform for event data, facilitating consistent training, evaluation, and optimization methods, thereby enabling reliable model comparison and performance benchmarking, as detailed in Section \ref{sec:experiment}.

\subsection{Reinforcement learning in soccer}
\label{ssec:related_rl_soccer}
The application of agent-based modeling in sports analytics has grown significantly, particularly for analyzing player behaviors and interactions. 
Data-driven analysis involves estimating various variables and functions (such as state variables, reward and policy functions, and action evaluation) in Markov decision processes (MDPs) from observed data representing a sub-problem of RL known as the inverse approach. In contrast, constructing an RL model to generate data in a virtual environment (i.e., the forward approach such as \cite{kurach2020google, fujii2024adaptive}) provides another way to understand the behavior of the modeled agents; however, the forward approach is beyond the scope of this study because it does not usually use the measured data. In this paper, we focus on introducing deep RL (DRL) based methods, these inverse approaches leverage large-scale data to evaluate actions based on their potential to maximize future rewards (for other approaches including the forward approach, see \cite{fujii2021data, fujii2025machine}).

In the context of on-ball actions for team sports, numerous studies have focused on estimating state-action value functions (Q-function) or other policy-related functions using soccer event data \cite{Liu2020,van2021leaving,van2023markov} and tracking data \cite{rahimian2021towards,rahimian2022beyond,rahimian2024towards}, as well as in sports like ice hockey \cite{Liu2018,Schulte2017}, badminton \cite{ding2022deep}, and basketball \cite{yanai2022q, chen2022reliable}. 
Despite these advancements, most models primarily represent teams using on-ball events, often neglecting evaluations of off-ball player dynamics at all time steps. To address this limitation, Nakahara et al. \cite{nakahara2023action} proposed a holistic RL framework capable of estimating Q-functions for multiple players, enabling simultaneous evaluation of both on- and off-ball players, even in the absence of explicit events. In this paper, we introduce the SAR format and RLearn package\footref{fn:rlearn} for the first time, which includes preprocessing, modeling, and visualization tools to improve accessibility and facilitate easier integration and analysis.
For other topics, research in inverse RL (IRL) has explored estimating reward functions directly from data \cite{Luo2020,rahimian2022inferring}. To model policies, trajectory predictions have been conducted using imitation learning \cite{Le17,le2017data,Teranishi2020,fujii2024decentralized} and behavioral modeling \cite{Zhan19,Yeh19,Li2021,fujii2024estimating}, which focus on mimicking observed behaviors through neural networks rather than optimizing them.

\section{OpenSTARLab}
\label{sec:openstarlab}

In this section, we present OpenSTARLab, a comprehensive framework for soccer analytics and research that integrates event annotation, data standardization, and predictive modeling capabilities. Section \ref{ssec:event_label_tool} introduces the STE Label Tool, a specialized software application for efficient soccer event annotation from match videos. Section \ref{ssec:UIED_Format} describes the Unified and Integrated Event Data (UIED) format, which standardizes event data across various providers such as GRF, StatsBomb, Wyscout, and DataStadium. Section \ref{ssec:pre-processing} details the Pre-processing package that facilitates the conversion of diverse data formats into UIED while maintaining data integrity and consistency. Section \ref{ssec:event} presents the Event Modeling package, which enables event prediction and analysis through state-of-the-art models, along with their practical applications. Finally, Section \label{ssec: sar} presents the RLearn Modeling package, which enables multi-agent decision-making analysis.The framework's limitations and future directions are discussed in Section \ref{ssec:limitations}.

The framework's complete workflow is illustrated in Figure \ref{fig:openstarlab}. The pipeline begins with raw soccer match videos, which are processed through the STE Label Tool for systematic event annotation. An intermediate step of pitch coordinate mapping may be required (see Section \ref{ssec:limitations} for details). The annotated data is then standardized into the UIED format using the OpenSTARLab Pre-processing Package. Finally, the OpenSTARLab Event Modeling Package enables comprehensive analysis through model training, inference, simulation, and various analytical applications such as heat maps and possession metrics.

\subsection{Spatial-Temporal Event (STE) label tool}
\label{ssec:event_label_tool}

\begin{figure}
\centering
\includegraphics[width=\textwidth]{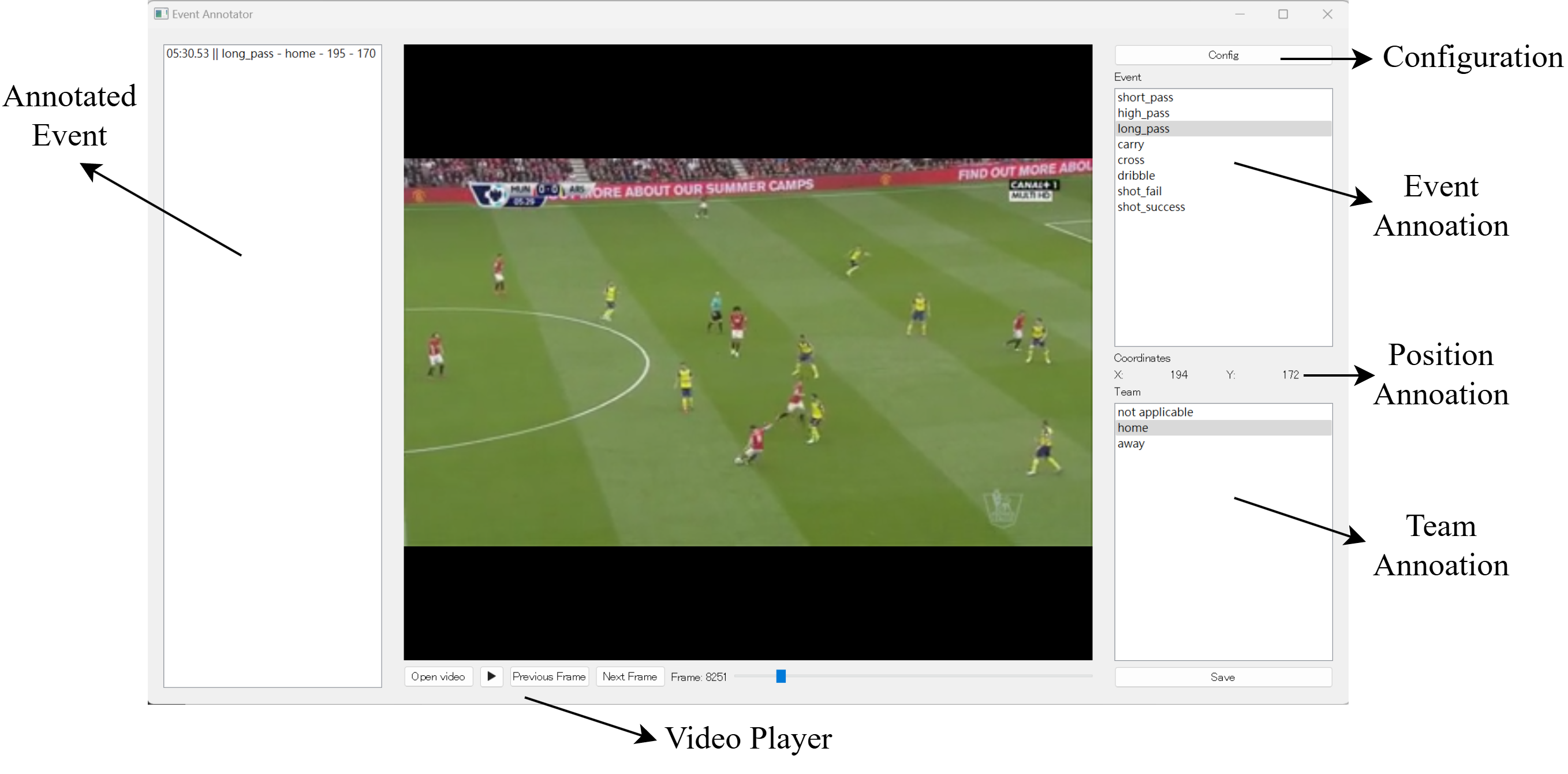}
\caption{Interface of STE label tool. Soccer match image from SoccerNet \citep{deliege2021soccernet}.}
\label{fig:interface}
\end{figure}

\begin{figure}
\centering
\includegraphics[width=0.5\textwidth]{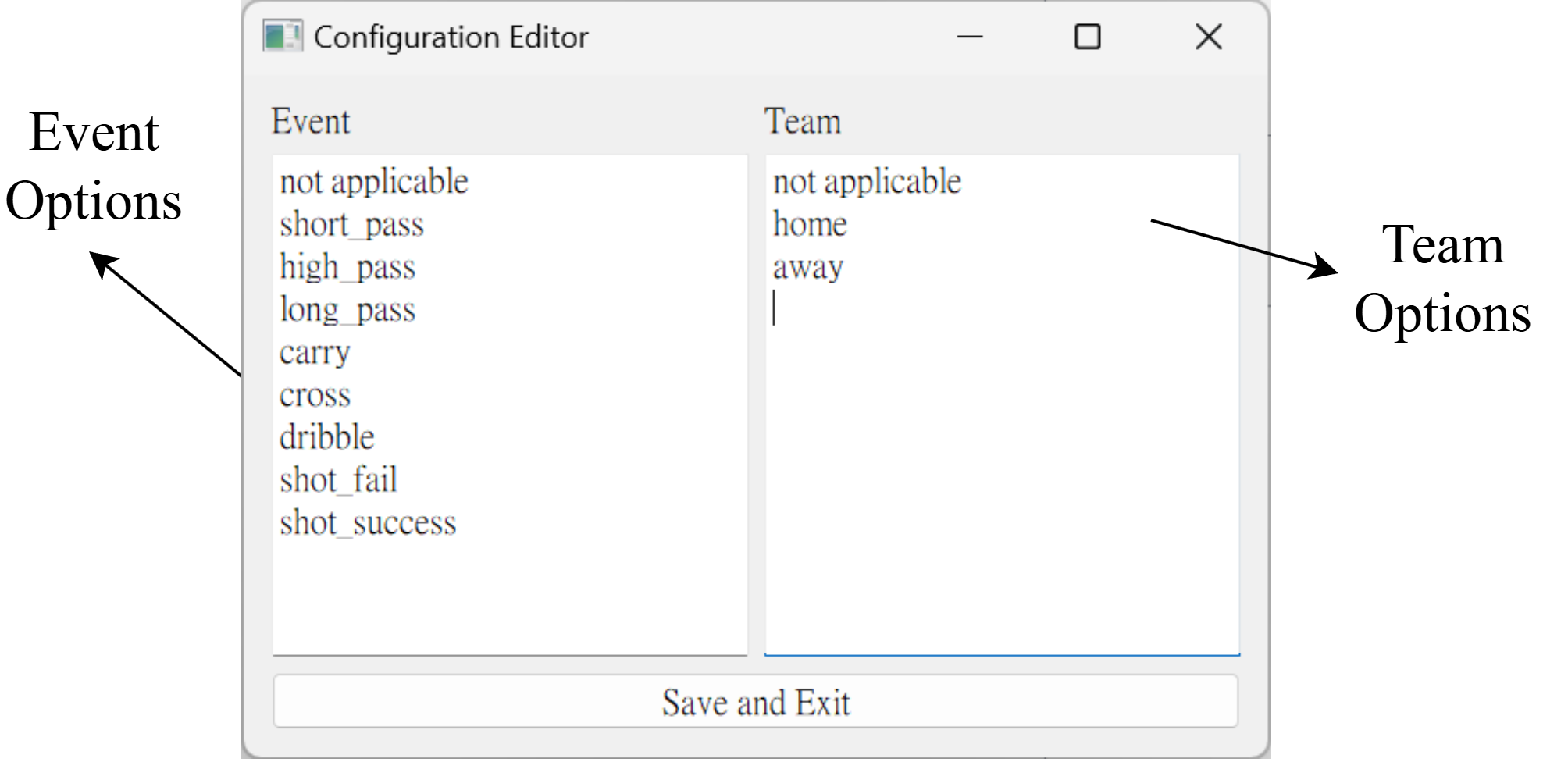}
\caption{Configuration of STE label tool.}
\label{fig:config}
\end{figure}

The STE Label Tool\footref{fn:event_label_tool} is a specialized software application inspired by the SoccerNet \citep{deliege2021soccernet} action annotation tool, designed to facilitate efficient and accurate annotation of soccer event data. While the original tool focused primarily on annotating isolated player action types, the STE Label Tool extends these capabilities to include broader event details such as event type, position, and team information. To enhance user intuitiveness and accessibility, the tool provides precise frame-by-frame navigation, displays the annotation workspace directly on the main interface, and allows users to edit annotation options directly on-screen. These enhancements make the tool particularly well-suited for labeling data critical to soccer analysis. It enables researchers and analysts to label key aspects of events in soccer match videos, which is essential for training machine learning models, especially in tasks related to predicting player actions, generating tactical insights, and evaluating player performance.

Developed using PyQt5, the STE Label Tool provides a flexible and user-friendly graphical interface, with an application layout focused on ease of use and efficient labeling workflows. The interface comprises three main columns, as shown in Figure \ref{fig:interface}:

\begin{itemize}
    \item \textbf{Left Column}: This section is dedicated to data input and displays previously annotated data (if any). Users can easily navigate through annotated events and corresponding frames, or remove annotations with minimal effort.
    
    \item \textbf{Middle Column}: This section enables frame-by-frame navigation, allowing users to precisely control time-based events. This feature is particularly useful for aligning annotations with exact moments in the video.
    
    \item \textbf{Right Column}: This serves as the primary workspace for annotating events. It displays options for configuration, event type selection, position selection (by simply clicking on the frame), and team selection. Additionally, the configuration section includes a pop-up window (Figure \ref{fig:config}) that allows users to edit options in a text editor-like format, where each row represents one customizable option.
\end{itemize}

This modular layout allows users to manage multiple functionalities without leaving the main screen, enhancing efficiency and minimizing task-switching. Event data collected using the tool are stored in formats compatible with the OpenStarLab packages, facilitating seamless integration with event prediction models.

The STE Label Tool is a critical component in the OpenStarLab pipeline for soccer analytics, enabling efficient data preprocessing and standardization for downstream model training and analysis. Its design and compatibility with OpenStarLab packages make it a valuable asset in advancing research in soccer event prediction and performance analysis.

\subsection{UIED format}
\label{ssec:UIED_Format}

\begin{table}[h!]
\centering
\begin{tabular}{lllll}
\toprule
\textbf{UIED} & \textbf{GRF} \citep{kurach2020google} & \textbf{Statsbomb} & \textbf{Wyscout} & \textbf{DataStadium} \\
\midrule
Short Pass & Short Pass & Ground Pass & Goal Kick & Direct FK Pass \\
                             &           & Low Pass    & Free Kick & Indirect FK    \\
                             &           & Half Start  & Simple Pass & KickOff       \\
                             &           &             & Hand Pass  & HomePass      \\
                             &           &             & Head Pass  & AwayPass      \\
                             &           &             & Smart Pass & PKPass        \\
                             &           &             & Throw In   & Through Pass  \\
                             &           &             &            & ThrowIn       \\
                             &           &             &            & Feed          \\
\hline
High Pass   & High Pass & High Pass   & High Pass & -              \\
\hline
Long Pass   & Long Pass & Ground Pass & Goal Kick & Direct FK Pass \\
                             &           & Low Pass    & Free Kick & Indirect FK    \\
                             &           &             & Simple Pass & KickOff      \\
                             &           &             &            & HomePass      \\
                             &           &             &            & AwayPass      \\
                             &           &             &            & PKPass        \\
                             &           &             &            & Through Pass  \\
                             &           &             &            & ThrowIn       \\
                             &           &             &            & Feed          \\
\hline
Shot        & Shot      & Shot        & Shot       & Shoot         \\
                             &           &             & Free Kick Shot & Direct FK Shot \\
\hline
Carry       & Sprint    & Carry       & Acceleration & -\\
\hline
Dribble     & Dribble   & Dribble     & Touch      & Dribble       \\
                             &           &             &            & Touch         \\
\hline
Cross       & -          & Cross       & Cross      & CK            \\
                             &           & Corner      & Free Kick Cross & Cross    \\
\botrule
\end{tabular}

\caption{Event type mappings across different soccer event data format. Short Pass and Long Pass are classified based on a pass length threshold of 45 meters. Additionally, special event markers—such as \texttt{end\_of\_possession} (denoted as ``\_''), \texttt{period\_over}, and \texttt{game\_over}—are appended at the end of each possession, period, and game to signal these transitions.}
\label{tab:event_type_mappings}
\end{table}

\begin{figure}
\centering
\includegraphics[width=0.65\textwidth]{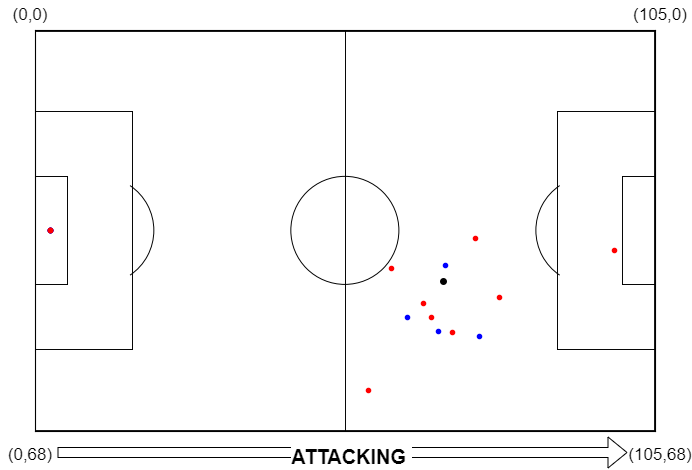}
\caption{UIED standardized pitch coordinates format. The red, blue, and black points represent the positions of the attacking team, defending team, and event location, respectively. The pitch is oriented such that the attacking direction is always from the left (\(x=0\)) to the right (\(x=105\)).}
\label{fig:pitch}
\end{figure}

\begin{table}[h!]
    \centering
    \begin{tabular}{llp{9.5cm}}
        \toprule
        \textbf{Variables} & \textbf{Type} & \textbf{Description} \\
        \midrule
        match\_id & int & Unique identifier for each match. \\
        
        poss\_id & int & Unique identifier for each possession within a match. \\
        
        team & str & The team associated with the event. \\
        
        home\_team & int & Indicator of whether the team is the home team (1: home, 0: away). \\
        
        \textbf{action} & str & Simplified and standardized description of the event action. \\
        
        success & int & Indicator of whether the event action was successful (1: success, 0: failure). \\
        
        goal & int & Indicator of whether the event resulted in a goal (1: goal, 0: no goal). \\
        
        home\_score & int & The current score of the home team. \\
        
        away\_score & int & The current score of the away team. \\
        
        goal\_diff & int & The goal difference (home\_score - away\_score). \\
        
        Period & int & The period of the match (1: 1st half, 2: 2nd half, etc.). \\
        
        Minute & int & The minute within the current period. \\
        
        Second & float & The second within the current minute. \\
        
        \textbf{seconds} & float & The total seconds elapsed since the start of the match, adjusted for different periods. (+15 minutes time for each period passed) \\
        
        \textbf{delta\_T} & float & The time difference between the current and previous events in seconds. \\
        
        \textbf{start\_x} & float & The x-coordinate of the event's starting location (scaled). \\
        
        \textbf{start\_y} & float & The y-coordinate of the event's starting location (scaled). \\
        
        deltaX & float & The change in the x-coordinate from the previous event. \\
        
        deltaY & float & The change in the y-coordinate from the previous event. \\
        
        distance & float & The distance covered by the event. \\
        
        dist2goal & float & The distance from the event's starting location to the center of the goal. \\
        
        angle2goal & float & The angle between the event's starting location and the goal, in radians. \\
        \botrule
    \end{tabular}
    \caption{Avalible variables for UIED format data. The bolded variables are the most basic and common variable in the event prediction model and performance analysis, as shown in Table \ref{Tab:compare_event_model}.}
    \label{tab:UIED_format}
\end{table}

The Unified and Integrated Event Data (UIED) format is designed to standardize event data across various soccer datasets, enabling seamless integration and comparative analysis. It provides a unified schema for encoding game events, allowing analysts and machine learning models to work with a consistent data format regardless of the source. (The distinction between UIED and existing formats is discussed in Section \ref{ssec:event_data_format_in_soccer}.) This standardization is crucial for accurate event prediction and performance evaluation across datasets that may have distinct naming conventions and structures.

To address differences across popular soccer datasets, the UIED format includes a comprehensive mapping of event types from datasets such as GRF, StatsBomb, Wyscout, and DataStadium. Table \ref{tab:event_type_mappings} illustrates these mappings for common event types, including passes, shots, and dribbles. For instance, a ``Short Pass'' in UIED corresponds to various terms across datasets, such as ``Ground Pass'' in StatsBomb and ``Simple Pass'' in Wyscout. By unifying these event types under a single nomenclature, UIED allows researchers and analysts to compare and aggregate events across sources without the need for manual re-labeling.

Additionally, UIED uses specific thresholds to differentiate event types. For example, passes are classified as either ``Short Pass'' or ``Long Pass'' based on a pass length threshold of 45 meters. Special event markers, such as \texttt{end\_of\_possession}, \texttt{period\_over}, and \texttt{game\_over}, are also included to signal key transitions in the match, facilitating timeline-based analysis.

Different soccer datasets tend to use different coordinate systems for event location. In the UIED format, the \texttt{start\_x} and \texttt{start\_y} coordinates represent the starting location of each event on the soccer pitch, standardized to a specific coordinate system for consistency. This coordinate system assumes a pitch width of 105 meters and a height of 68 meters, following FIFA's (Fédération Internationale de Football Association) recommended field dimensions. The coordinates are scaled accordingly to fit within this layout, as illustrated in Figure \ref{fig:pitch}.

The origin point, \((0, 0)\), is defined at the top-left corner of the pitch, with:
\begin{itemize}
    \item The \textbf{x-axis} increasing horizontally from left to right, spanning from \( x = 0 \) (left end) to \( x = 105 \) (right end).
    \item The \textbf{y-axis} increasing vertically from top to bottom, spanning from \( y = 0 \) (top end) to \( y = 68 \) (bottom end).
\end{itemize}

This setup facilitates consistent analysis of event locations, player positioning, and ball movement across datasets. For example:
\begin{itemize}
    \item The center of the pitch is located at approximately \( (52.5, 34) \).
    \item The \textbf{goal areas} are positioned near \( x = 0 \) and \( x = 105 \), with events closer to these ends indicating attacking or defensive plays, depending on the team's perspective.
    \item Events in the \textbf{attacking direction} flow from the left to the right end of the pitch, as indicated by the ``ATTACKING'' arrow at the bottom of Figure \ref{fig:pitch}.
\end{itemize}

This standardized system simplifies the evaluation of each event's spatial context, such as distance to the goal (calculated from \texttt{start\_x} and \texttt{start\_y} to the goal's coordinates) or the angle relative to the goal, which are essential metrics in shot analysis and predictive modeling.

Finally, the UIED format defines a comprehensive set of variables that capture essential aspects of each event. Table \ref{tab:UIED_format} provides an overview of these variables. Key variables in the UIED format, such as \texttt{action}, \texttt{seconds}, \texttt{delta\_T}, \texttt{start\_x}, and \texttt{start\_y}, and derived variables are designed to support in-depth analysis of gameplay and facilitate robust model training. This standardized structure allows for precise evaluations of gameplay dynamics and player performance across datasets, forming a reliable foundation for advanced soccer analytics.

\subsection{SAR Format}
\label{ssec: SAR_Format}

\begin{table}[h!]
    \centering
    \begin{tabular}{llp{9cm}}
        \toprule
        \textbf{Variables} & \textbf{Type} & \textbf{Description} \\
        \midrule
        match\_id & int & Unique identifier for each match. \\
        
        frame\_id & int & Unique identifier for each frame within a match. \\
        
        team & str &  The team associated with the event. \\
        
        team\_id & int & Unique identifier for each team. \\ 
        
        home\_team & int & Indicator of whether the team is the home team ($1$: home, $0$: away). \\

        player\_name & str & The player associated with the event. \\

        \textbf{player\_id} & int & Unique identifier for each player. \\
        
        jersey\_number & int & The player jersey\_number associated with the event. \\
        
        player\_role\_id & int & The Identifier for each position of each player ($1$: GK, $2$: DF, $3$: MF, $4$: FW). \\
        
        action & str & Simplified and standardized description of the event action. \\
        
        \textbf{action\_id} & int & Unique identifier for each action. \\ 

        success & int & Indicator of whether the event action was successful ($1$ for success, $0$ for failure). \\
        
        is\_goal & int & Indicator of whether the event resulted in a goal ($1$: goal, $0$: no goal). \\
        
        is\_shot & int & Indicator of whether the event is a shot ($1$: shot, $0$: no shot). \\
        
        is\_pass & int & Indicator of whether the event is a pass ($1$: pass, $0$: no pass). \\
        
        is\_dribble & int & Indicator of whether the event is a dribble ($1$:dribble, $0$: no dribble). \\
        
        is\_cross & int & Indicator of whether the event is a cross ($1$: cross, $0$: no cross). \\
        
        is\_through\_pass & int & Indicator of whether the event is a through\_pass ($1$: through\_pass, $0$: no through\_pass). \\
        
        is\_ball\_recovery & int & Indicator of whether the event is a recovery ($1$: recovery, $0$: no recovery). \\
        
        is\_block & int & Indicator of whether the event is a block ($1$: block, $0$: no block). \\
        
        is\_clearance & int & Indicator of whether the event is a clearance ($1$: clearance, $0$: no clearance). \\
        
        is\_interception & int & Indicator of whether the event is a interception ($1$: interception, $0$: no interception). \\
        
        Period & int & The period of the match ($1$: 1st half, $2$: 2nd half, etc.). \\
        
        seconds & float & The total seconds elapsed since the start of the match, adjusted for different periods. \\        
        
        start\_x & float & The x-coordinate of the player location when event's starting (scaled). \\
        
        start\_y & float & The y-coordinate of the player location when event's starting (scaled). \\

        ball\_x & float & The x-coordinate of the ball location when event's starting (scaled). \\
        
        ball\_y & float & The y-coordinate of the ball location when event's starting (scaled). \\
        
        ball\_touch & int & Only event valid as play $1$ Ball out, fouls, etc. $0$. \\
        
        series\_num & int & Sequential number of the sequence of in-play of the match. \\
        
        history\_num & int & No. of the history in chronological order of the play of the match (the start of the match is $1$, and after that, the number is $+1$ or higher than the previous history No.). \\
        
        \textbf{attack\_history\_num} & int & Number for a single series of attacks. \\
        
        attack\_start\_num & int & First history No. within the same attack\_history\_num. \\
        
        attack\_end\_history\_num & int & Last history No. within the same attack\_history\_num. \\
        
        \botrule
    \end{tabular}
    \caption{Avalible variables for SAR format event data. The bolded variables are the most essential variable in the reward prediction model and performance analysis.}
    \label{tab:SAR_event_format}
\end{table}

\begin{table}[h!]
    \centering
    \begin{tabular}{llp{9cm}}
        \toprule
        \textbf{Variables} & \textbf{Type} & \textbf{Description} \\
        \midrule
        match\_id & int & Unique identifier for each match. \\
        
        frame\_id & int & Unique identifier for each frame within a match. \\
        
        home\_team & int & Indicator of whether the team is the home team ($1$: home, $0$: away). \\
        
        jersey\_number\_id & int & The player jersey\_number associated with the event. \\
        
        \textbf{x} & float & The x-coordinate of the player location scale by the field size. \\
        
        \textbf{y} & float & The y-coordinate of the player location scale by the field size. \\

        \botrule
    \end{tabular}
    \caption{Avalible variables for SAR format tracking data. The bolded variables are the most essential variable in the reward prediction model and performance analysis.}
    \label{tab:SAR_tracking_format}
\end{table}

The State-Action-Reward (SAR) format standardizes event and tracking data across diverse soccer datasets, facilitating seamless integration and comparative analysis. Similar to the UIED format, this standardization is essential for accurate reward prediction and performance evaluation, especially when datasets use different naming conventions and structures.

The SAR format is available in datasets such as StatsBomb, SkillCorner, and DataStadium. While it maintains the same pitch width and height as the UIED format (Figure \ref{fig:pitch}), it also aligns with the UIED format in both structure and direction of attack, which progresses from left to right on the pitch (Figure \ref{fig:pitch}). However, its coordinate system differs, with the origin point positioned at the center of the pitch. The coordinate system in the SAR format is defined as follows:  

\begin{itemize}
    \item The \textbf{x-axis} increases horizontally from left to right, ranging from \( x = -52.5 \) (left end) to \( x = 52.5 \) (right end).
    \item The \textbf{y-axis} increases vertically from top to bottom, ranging from \( y = -34 \) (top end) to \( y = 34 \) (bottom end).
\end{itemize}  

For tracking data, the velocity and acceleration of all players and the ball were calculated, with missing data supplemented where necessary. Event and tracking data were synchronized based on frame\_id. Additionally, attack sequences were segmented using ``attack\_history\_num'', following the approach of a previous study \citep{nakahara2023action}. By structuring data into distinct attack sequences, rather than treating an entire match as a single sequence, this method enables more granular learning and evaluation of offensive plays. To ensure consistency, only sequences with a minimum of 50 frames and a maximum of 300 frames were included in the dataset.

Finally, the SAR format defines a comprehensive set of variables capturing key aspects of each event. Table \ref{tab:SAR_event_format} provides an overview of the event data variables, while Table \ref{tab:SAR_tracking_format} details the tracking data variables. Key variables such as \texttt{player\_id}, \texttt{action\_id}, \texttt{attack\_series\_num}, \texttt{x}, and \texttt{y}, along with derived variables, are designed to support in-depth gameplay analysis and enhance model training.

\subsection{Pre-processing package}
\label{ssec:pre-processing}
The OpenStarLab Pre-processing package\footref{fn:preprocessing} is a comprehensive tool designed to standardize and streamline event data and tracking data in soccer across various providers. It allows users to easily convert raw event data from different formats into the Unified and Integrated Event Data (UIED) format or other formats used in event predictive studies \citep{simpson2022seq2event,yeung2023events,yeung2023transformer,yeung2024unveiling,mendes2024towards,mendes2024forecasting}, and  State-Action-Reward (SAR) format in reinforcement learning, providing a cohesive and standardized dataset for analysis.

\subsubsection{Key features}
The key features of OpenSTARLab Pre-processing package are as follows:
\begin{itemize}
    \item \textbf{Multi-Provider Compatibility}: The package supports multiple data providers, including DataFactory, DataStadium, Metrica, Opta, Robocup 2D Simulation, SoccerTrackv2 (BePro), Sportec, Statsbomb, Statsbomb with Skillcorner Tracking Data, and Wyscout, allowing seamless integration regardless of the original data format. For consistency, most providers' (see Section \ref{ssec:limitations}) data structures could be mapped to the UIED standard or the formats used in previous studies\footnote{Details of other formats: \url{https://openstarlab.readthedocs.io/en/latest/Pre_Processing/Sports/Event_data/Data_Format/Football/other.html}}.

    \item \textbf{Automated Preprocessing Pipeline}: With functions tailored to each provider, the package can handle large volumes of data by loading, parsing, and transforming it into the UIED format. Users specify parameters such as file paths and IDs, and the package handles the necessary transformations, including coordinate standardization, event categorization, and temporal adjustments.

    \item \textbf{Scalable Processing}: The package is built with scalability in mind, supporting parallel processing through the \texttt{max\_workers} parameter. This allows users to customize the level of parallelization based on available computing resources, optimizing performance when dealing with large datasets.

    \item \textbf{UIED-Standardized Outputs}: The resulting data is output in the UIED format (detailed in Section \ref{ssec:UIED_Format}), with standardized columns for match and possession IDs, event actions, spatial information, and game statistics. This uniform structure simplifies subsequent analysis and model training tasks by providing consistent data regardless of the source.
    
    \item \textbf{SAR Format Option}:
    The package also supports the SAR format \ref{ssec: SAR_Format} specifically tailored for deep learning tasks (in particular DRL by conputing state, action, and reward).
    By aligning event and tracking data at each time frame within a unified framework, it enables seamless integration and supports advanced deep learning applications.
\end{itemize}

\subsubsection{Customization and flexibility}

The OpenStarLab Pre-processing package allows users to specify provider-specific parameters and preprocessing options, offering flexibility for different project requirements. Its modular structure enables easy integration into larger workflows, while broad compatibility with various file formats ensures minimal setup effort.

Each provider’s function parameters can be customized to match specific file paths or other requirements, producing standardized output ready for UIED-based analysis. For further details, refer to the documentation of OpenStarLab\footnote{\label{temp}Comprehensive documentation on UIED data providers: \url{https://openstarlab.readthedocs.io/en/latest/Pre_Processing/Sports/Event_data/Data_Provider/Football/index.html}}. Below is an example demonstrating how to use the OpenStarLab Pre-processing package to convert Wyscout data \citep{pappalardo2019public} into the UIED format.

\begin{tcolorbox}[colback=gray!5!white, colframe=gray!75!black, title=Python code for Example usage]
\begin{verbatim}
from preprocessing import Event_data

wyscout_df = Event_data(data_provider="wyscout",
                        event_path="path/to/event/folder",
                        match_folder="path/to/match/folder",
                        preprocess_method="UIED",
                        max_workers=1).preprocessing()
\end{verbatim}
\end{tcolorbox}

Similar to the UIED-based analysis, the SAR-based analysis enables the customization of function parameters for each provider to accommodate specific file paths and other requirements, ultimately generating standardized output suitable for SAR-based analysis. For more details, please refer to the OpenStarLab documentation\footnote{Comprehensive documentation on SAR data providers \url{https://openstarlab.readthedocs.io/en/latest/Pre_Processing/Sports/SAR_data/Data_Provider/index.html}}. Below is an example demonstrating how to use the OpenStarLab Pre-processing package to convert DataStadium data \citep{pappalardo2019public} into the SAR format.

\begin{tcolorbox}[colback=gray!5!white, colframe=gray!75!black, title=Python code for SAR Example usage]
\begin{verbatim}
from preprocessing import SAR_data

datastadium_df = SAR_data(data_provider="datastadium",
                        event_path="path/to/event/folder",
                        tracking_h_path="path/to/tracking_h/folder,
                        tracking_a_path="path/to/tracking_a/folder,
                        max_workers=1).preprocessing()
\end{verbatim}
\end{tcolorbox}

In summary, this package provides a powerful and efficient tool for soccer event and tracking data preprocessing, transforming disparate data sources into a standardized UIED format and SAR format for streamlined analysis and application in machine learning models.

\subsection{Event modeling package}
\label{ssec:event}
The OpenSTARLab Event package\footref{fn:event} is an essential toolkit for event prediction-based modeling, offering users streamlined functionalities for training, inference, simulation, and applications in the sports domain. This package is optimized to work with data formatted by the OpenSTARLab PreProcessing package, facilitating smooth integration and ensuring a consistent workflow across OpenSTARLab repositories.

\subsubsection{Key functionalities}
The Event package provides comprehensive support for several event prediction models tailored for soccer analytics, including Seq2Event \citep{simpson2022seq2event}, NMSTPP \citep{yeung2023transformer}, LEM \citep{mendes2024towards}, FMS \citep{baron2024foundation}, and the Majority Class (MAJ) model, a simple baseline that predicts the most frequent event class from the training data. In addition, the package offers a variety of functions that optimize the event modeling workflow. Key features include:

\begin{enumerate}
 
    \item \textbf{Data Loading and Preprocessing Functions for UIED Format}: These functions facilitate the loading, cleaning, and organization of UIED data, ensuring seamless integration with new model architectures. They provide researchers and analysts with an efficient foundation for developing and testing event prediction models.

    \item \textbf{Training and Hyperparameter Tuning Functions}: The package includes built-in training functions that support hyperparameter optimization using Optuna \citep{optuna_2019}, one of the most widely used frameworks for hyperparameter tuning. For convenience, hyperparameter configurations can be easily specified in a simple YAML file, minimizing user effort. More details are provided in Appendix \ref{app:example_event}.

    \item \textbf{Inference and Simulation Functions}: Apart from traditional inference functions, the package provides simulation functions to simulate events with greedy selection (select the one with the highest probability) or probabilistic selection, supporting the application of counterfactual analysis. For example, in FMS \citep{baron2024foundation}, researchers can simulate how different actions or player movements might impact the game’s outcome.

    \item \textbf{Evaluation Metrics}: The package includes built-in functions to track and display key metrics such as loss function value, accuracy, F1-score, mean absolute error (MAE), and cross-entropy loss during training, inference, and simulation. These metrics are crucial for assessing model reliability and precision, enabling cross-model comparisons and a detailed evaluation of simulation performance, which was previously unavailable. More details are provided in Section \ref{sec:experiment}.

\end{enumerate}

\subsubsection{Application of the event model in Soccer}
Event prediction models have a wide application in soccer analytics, particularly for generating visualizations and metrics that capture spatial patterns, possession efficiency, and overall team performance. These Event Model applications enable a deep understanding of soccer gameplay dynamics, supporting data-driven decisions and performance evaluations tailored to the nuances of soccer. As a result, the package encompasses the most important applications, with available functionalities including\footnote{For more details on Event Model applications in soccer analytics: \url{https://openstarlab.readthedocs.io/en/latest/Event_Modeling/Sports/Soccer/application.html}}:

\begin{itemize}

\item \textbf{Heat Map of the Predicted Next Event Location} \citep{mendes2024forecasting}:
The event model allows analysts to generate heat maps predicting the location of subsequent events. By visualizing the probability distribution of the next event, this tool provides valuable insights into potential hotspots for player actions, helping teams to analyze spatial trends and refine strategies. For example, in defensive planning, understanding likely zones for opponent actions can guide positioning and movement planning. Similarly, offensive analyses can reveal opportunities to exploit open spaces based on opponent tendencies.

\item \textbf{Holistic Possession Utilization Score (HPUS) and HPUS+} \citep{yeung2023transformer}:
The HPUS and HPUS+ metrics quantify possession effectiveness over time. HPUS captures the overall value of each possession within a match, while HPUS+ refines this assessment to reflect the realization of that possession value. These metrics are particularly useful for tracking team strategy across intervals, as they highlight periods of high or low possession productivity. For instance, a steady HPUS score may indicate a team’s control over the game, while a fluctuating HPUS+ score could suggest varied effectiveness in creating goal-scoring opportunities.

\ \ Through segment-based visualization, HPUS and HPUS+ illustrate how teams capitalize on their possessions across the match duration, offering insights into momentum shifts, defensive or offensive efficiency, and situational adjustments. These metrics are effective tools for both in-match tactical evaluations and post-match performance summaries.

\item \textbf{Possession Utilization Score (Poss-Util) and Poss-Util+} \citep{simpson2022seq2event}:
The Poss-Util and Poss-Util+, enhance performance insights by measuring a team's success in converting possession into scoring chances. Poss-Util highlights a team's efficiency by comparing their possession time to meaningful actions, helping to gauge each team’s productivity and strategic approach. An elevated Poss-Util score typically indicates efficient use of ball control to generate offensive opportunities, whereas a low score might suggest prolonged but unproductive possession phases.

\ \ This metric is also applicable to individual and comparative analyses, allowing analysts to track performance across multiple teams. By observing Poss-Util scores over a season, coaches can evaluate tactical changes, assess opponent weaknesses, and adjust training focus to optimize possession strategies.

\end{itemize}

In conclusion, the OpenSTARLab Event package is a comprehensive tool for event modeling in soccer, designed to facilitate the analysis and prediction of key in-game events. It supports various models, enabling users to generate visualizations such as heat maps and calculate advanced metrics. These metrics provide valuable insights into team performance, possession efficiency, and tactical strategies throughout a match. The package is a powerful resource for both analysts and researchers seeking to enhance their understanding of soccer gameplay through data-driven methodologies.

\subsection{Reinforcement learning modeling (RLearn) package}
\label{ssec: sar}

The OpenSTARLab RLearn package\footref{fn:rlearn} is a toolkit designed for reinforcement learning (RL) in multi-agent decision-making models, operating at each timestep. Similar to the Event Modeling package, RLearn is compatible with data formatted by the OpenSTARLab PreProcessing package, ensuring a streamlined and consistent workflow across OpenSTARLab repositories.

While RL models in soccer are diverse, the RLearn package focuses on multi-player RL at each timestep \cite{nakahara2023action}. In contrast, many other RL models—such as event-driven single-agent models \cite{Liu2020,van2021leaving,van2023markov,rahimian2021towards,rahimian2022beyond,rahimian2024towards}—are very similar to those in Event Modeling package, provided the data and output is formatted appropriately. In the context of imitation learning \cite{Le17, fujii2024decentralized} and behavioral modeling \cite{Zhan19, Yeh19, Li2021, fujii2024estimating}, these models solely utilize tracking data, and disregard the event data.

For these reasons, the RLearn package is particularly valuable in its ability to handle both event and tracking data of multi-player at each timestep, offering a comprehensive solution for RL in soccer. Key features of the RLearn package include:

\begin{enumerate}
    \item \textbf{Data Loading and Preprocessing Functions for SAR Format}: These functions, similar to those for the UIED format in event modeling, facilitate the loading, cleaning, and organization of SAR data (see Section \ref{ssec: SAR_Format}). This ensures seamless integration with new model architectures and facilitates data preparation for RL tasks.
    
    \item \textbf{Modeling}: The provided baseline models include Multi-Layer Perceptron (MLP), Gated Recurrent Unit (GRU), and Long Short Term Memory (LSTM). While previous study \citep{nakahara2023action} has focused solely on GRU, this study further examines the effectiveness of MLP or LSTM to enable a comparative analysis of whether the use of time-series data improves performance. 
    
    \item \textbf{Evaluation Metrics}: The package includes a built-in functionality to monitor key performance indicators, such as loss function values and action prediction accuracy, both during training and inference. The action prediction accuracy was evaluated using TorchMetrics, a library provided by PyTorch, to assess the model’s classification performance across $16$ action types. Following \citep{nakahara2023action}, the Temporal Difference (TD) error was adopted as a loss function to evaluate the validity of the Q-values learned by the model.
    
    \item \textbf{Visualization Function}: The package offers a visualization tool that enables users to observe evaluation metrics for continuous actions of multiple agents. This allows for quantitative evaluation alongside expert qualitative assessments, by visualizing the effectiveness of players' actions in real time.
\end{enumerate}

\subsection{Limitations}
\label{ssec:limitations}

For event labeling tools, when using broadcast video, the annotated coordinates may not be directly convertible to pitch coordinates with the OpenStarLab packages. However, many existing studies \citep{chen2019sports,sha2020end,chu2022sports,theiner2023tvcalib,gutierrez2024no} and repositories provide methods for such conversions, offering alternative solutions. For a fixed camera angle, pitch coordinate mapping can be achieved using simple homography estimation, as demonstrated in \citep{scott2022soccertrack}.

Regarding the UIED format, while the preprocessing package supports loading data from most major soccer data providers to generate a standardized event data version, full mapping to the UIED format is currently only supported for GRF \citep{kurach2020google}, StatsBomb, Wyscout, and DataStadium. For instance, the RoboCup 2D data includes only pass events, making it impossible to categorize other types of events, such as shots and carry.

As for the event package, although significant effort has been made to optimize the code and models, performance may still be influenced by the specific version of PyTorch being used. Additionally, while the codebase is standardized, there is still room for further optimization. Despite these factors, the results should remain consistent and provide a fair basis for comparison across different setups.

\section{Experiments and Results}
\label{sec:experiment}

In this section, the experimental setup and results for evaluating the proposed framework for soccer match data were presented. Section \ref{ssec:dataset} introduces the datasets used in the event modeling experiments, including details on the Wyscout and StatsBomb datasets, which serve as the foundation for model training and evaluation. Section \ref{ssec:preprocessing_and_configuration} outlines the preprocessing and configuration steps, including data preparation, model setup, and hyperparameter tuning. Section \ref{ssec:comparison_of_model} provides a detailed comparison of the performance of different models, assessing error metrics for event predictions. Section \ref{ssec:event_model_sim_performance} discusses the simulation performance of the best-performing model, focusing on event count, action classification, and predictive accuracy for spatiotemporal coordinates at each timestep. 
Sections \ref{ssec:rl_dataset}, \ref{ssec:rl_details}, and \ref{ssec:rl_performance} described the RL experiments using RLearn package. 
Finally, Section \ref{ssec:application_on_soccer_analytics} demonstrates the application of the event model with real-world soccer analytics.

\subsection{Dataset in Event Modeling}
\label{ssec:dataset}
In the event modeling experiments, two event datasets were used, each capturing actions performed by players in control of the soccer ball. These datasets document a variety of action types, such as passes, shots, fouls, and more. For each recorded action, additional details are provided, including the position on the soccer pitch (in (x, y) coordinates), the time of the event, and the outcome of the action. More detailed information about the datasets is provided below:

\textbf{Wyscout dataset}: It includes soccer match event data from the 2017/2018 season across the top five leagues: the Premier League, La Liga, Ligue 1, Serie A, and Bundesliga. This dataset was obtained from the WyScout Open Access Dataset \citep{pappalardo2019public} and is currently recognized as the largest publicly available collection of soccer match event data, aimed at advancing research in soccer data analytics. For computational efficiency, the Premier League was specifically chosen due to its high-intensity gameplay, which leads to greater variation between events.

\textbf{StatsBomb Dataset}: To test the models on the latest data, the 2023/2024 La Liga season dataset was purchased from StatsBomb, as La Liga teams frequently compete at the highest level, including winning the Champions League against teams from other major soccer leagues in the 2023/2024 season. However, due to terms and conditions, this dataset could not be published as part of this study. It includes all 380 matches played by the 20 La Liga teams.

\subsection{Preprocessing and Configuration in Event Modeling}
\label{ssec:preprocessing_and_configuration}
The two datasets in Section \ref{ssec:dataset} were split into training, validation, and test sets with a 60/20/20 ratio. The data were then converted to the UIED format (see Section \ref{ssec:UIED_Format}) using the OpenSTARLab Pre-processing package (see Section \ref{ssec:pre-processing}). Afterward, all models were trained with the OpenSTARLab Event package (see Section \ref{ssec:event}).

For each model, hyperparameter tuning was conducted with a maximum of 100 trials, and training was performed on an NVIDIA GeForce RTX 3090 GPU, with training for a single model taking approximately three to four days. Detailed configuration settings, training code, and pre-trained model parameters are available in the OpenSTARLab documentation\footnote{OpenSTARLab document on model training: \url{https://openstarlab.readthedocs.io/en/latest/Event_Modeling/Sports/Soccer/model.html}}.

The prediction task for each event includes determining the action type, time, and (x, y) coordinates. To evaluate performance, the following metrics were utilized: accuracy and F1-score for action type prediction, when accounting for the class imbalance in action types, F1-score provides a more accurate reflection of performance. Mean Absolute Error (MAE) was used to assess the predictions of time and spatial coordinates, which offer an intuitive understanding of model accuracy within the context of a football match. In addition, the Floating-point operations per second (FLOPs) and number of parameters (Num Prams) were included to show the computational efficiency and model complexity of each approach, providing insights into their scalability and resource requirements when applied to large-scale soccer datasets.

The models evaluated in this study include the baseline majority class model (MAJ), Seq2Event \citep{simpson2022seq2event}, NMSTPP \citep{yeung2023transformer}, LEM\_1 \citep{mendes2024towards}, LEM\_3 \citep{mendes2024forecasting}, and FMS \citep{baron2024foundation}. The key distinction between LEM\_1 and LEM\_3 lies in their model input: LEM\_1 uses only the most recent event, whereas LEM\_3 leverages events from the three most recent timesteps. Finally, NMSTPP+360 \citep{yeung2023events,yeung2024unveiling} was excluded from this evaluation due to its dependency on player coordinate data, which is not available in any current large public soccer datasets.

\subsection{Comparison of event model performance}
\label{ssec:comparison_of_model}
\begin{table}[]
\centering
\begin{tabular}{llllllll}
\toprule
\parbox{3cm}{ \textbf{Model} \\ \textbf{(Publication Year)}} &\parbox{1.00cm}{ \textbf{Action} \\ \textbf{Acc.}}$\uparrow$  & \parbox{1.00cm}{ \textbf{Action} \\ \textbf{F1}}$\uparrow$ & \parbox{1cm}{ \textbf{Time} \\ \textbf{-MAE}}$\downarrow$ & \textbf{X-MAE} $\downarrow$ & \textbf{Y-MAE} $\downarrow$ & \textbf{FLOPs} & \parbox{1.0cm}{ \textbf{Num} \\ \textbf{Prams}} \\ 
\midrule
\textbf{Wyscout Dataset} &&&&&&\\
MAJ       & 0.57                          & 0.08                         & 3.60                     & 18.97                    & 52.55                    & \multicolumn{1}{l}{-}     & \multicolumn{1}{l}{-}           \\
Seq2Event (2022) \citep{simpson2022seq2event} & 0.67                          & 0.16                         & 3.41                     & 7.11                     & 15.72                    & 112M                 & 135K                         \\
NMSTPP (2023) \citep{yeung2023transformer}    & 0.67                           & 0.17                         & 3.34                       & \textbf{6.94}            & \textbf{15.08}           & 296M                 & 121K                          \\
LEM\_1 (2024) \citep{mendes2024towards}   & \textbf{0.67}                 & 0.17                         & 3.07                     & 8.34                    & 21.44                    & 50M                  & 98K                           \\
LEM\_3 (2024) \citep{mendes2024forecasting}   & 0.67                          & \textbf{0.20}                & \textbf{2.69}            & 7.62                     & 21.83                    & 20M                  & 39K     \\
FMS (2024) \citep{baron2024foundation}     & 0.67                          & 0.16                         & 3.27                     & 11.27                   & 24.19                    & 930M                 & 782K                          \\
\textbf{StatsBomb Dataset} &&&&&&\\
MAJ       & 0.40          & 0.06          & 2.76          & 20.72         & 33.32         & -          & -           \\ 
Seq2Event (2022) \citep{simpson2022seq2event} & 0.65          & 0.23          & 2.43          & 7.22          & 6.86          & 4.03B      & 413K     \\ 
NMSTPP (2023) \citep{yeung2023transformer}    & 0.65          & 0.23          & 2.53          & 7.38          & \textbf{6.86} & 2.02B      & 217K     \\ 
LEM\_1 (2024) \citep{mendes2024towards}    & 0.65          & 0.24          & 2.23          & 7.36          & 8.21          & 66M      & 128K    \\ 
LEM\_3 (2024) \citep{mendes2024forecasting}     & \textbf{0.66} & \textbf{0.25} & \textbf{2.07} & \textbf{7.07} & 8.32          & 19M      & 38K     \\ 
FMS (2024) \citep{baron2024foundation}       & 0.65          & 0.24          & 2.35          & 7.77          & 8.82          & 3.66B      & 1.29M   \\ 

\botrule
\end{tabular}
\caption{Comparison of model performance on soccer event prediction. Arrows indicate whether a higher ($\uparrow$) or lower ($\downarrow$) value is better for each metric. Notation: B = billion, M = million, K = thousand.  The models are ranked by publication year. The bold value indicates the performance, determined by the unrounded value.}
\label{tab:comparison_of_model}
\end{table}

This section compared the performance of the existing event model with the Wyscout and StatsBomb datasets, as detailed in Section \ref{ssec:dataset}, using the preprocessing and configuration outlined in Section \ref{ssec:preprocessing_and_configuration}. The comparison provided a benchmark for future studies. A summary of the performance of various models on soccer event prediction was presented in Table \ref{tab:comparison_of_model}.

First, considering \textbf{Action Accuracy and F1-Score}, the LEM\_3 model achieved the highest performance for both metrics on both datasets. This trend indicated the effectiveness of LEM\_3 in handling soccer event prediction tasks, likely due to its use of three previous timestep events, which provided more contextual information compared to the LEM\_1 model, which only utilized one timestep. Additionally, the LEM\_3 model's architecture proved more effective, as models like FMS, Seq2event, and NMSTPP, which used 9, 40, and 40 previous events respectively, also contributed to its superior performance. Overall, the accuracy of all models was similar, but the LEM models outperformed the others in terms of the F1-score, demonstrating better handling of imbalanced action classes.

In terms of \textbf{Time and Location MAE}, LEM\_3 also demonstrated superior performance in predicting the timing across both datasets and the spatial x-coordinate of events in the StatsBomb dataset. The NMSTPP model performed the best in predicting the xy coordinates but lagged behind LEM\_3 in the x-coordinate prediction on the StatsBomb dataset. When considering \textbf{Computational Efficiency} in terms of FLOPs and Parameter Count, the LEM models, particularly LEM\_3, were notable for their efficiency, with lower FLOPs and parameter counts compared to other models, which could require billions of FLOPs. LEM\_3's efficiency highlighted its strong performance without excessive computational overhead, making it a practical choice for large-scale implementations and simulations.

Finally, a \textbf{Dataset-Specific Performance} trend emerged, where models generally achieved better action accuracy and lower MAE values on the StatsBomb dataset compared to Wyscout. This difference suggested that the seasons and the league of choice could influence the performance of the event prediction task. Overall, the LEM\_3 model demonstrated the best balance between performance and computational efficiency across both datasets.

\subsection{Event Model simulation performance}
\label{ssec:event_model_sim_performance}
\begin{figure}
\centering
\includegraphics[width=\textwidth]{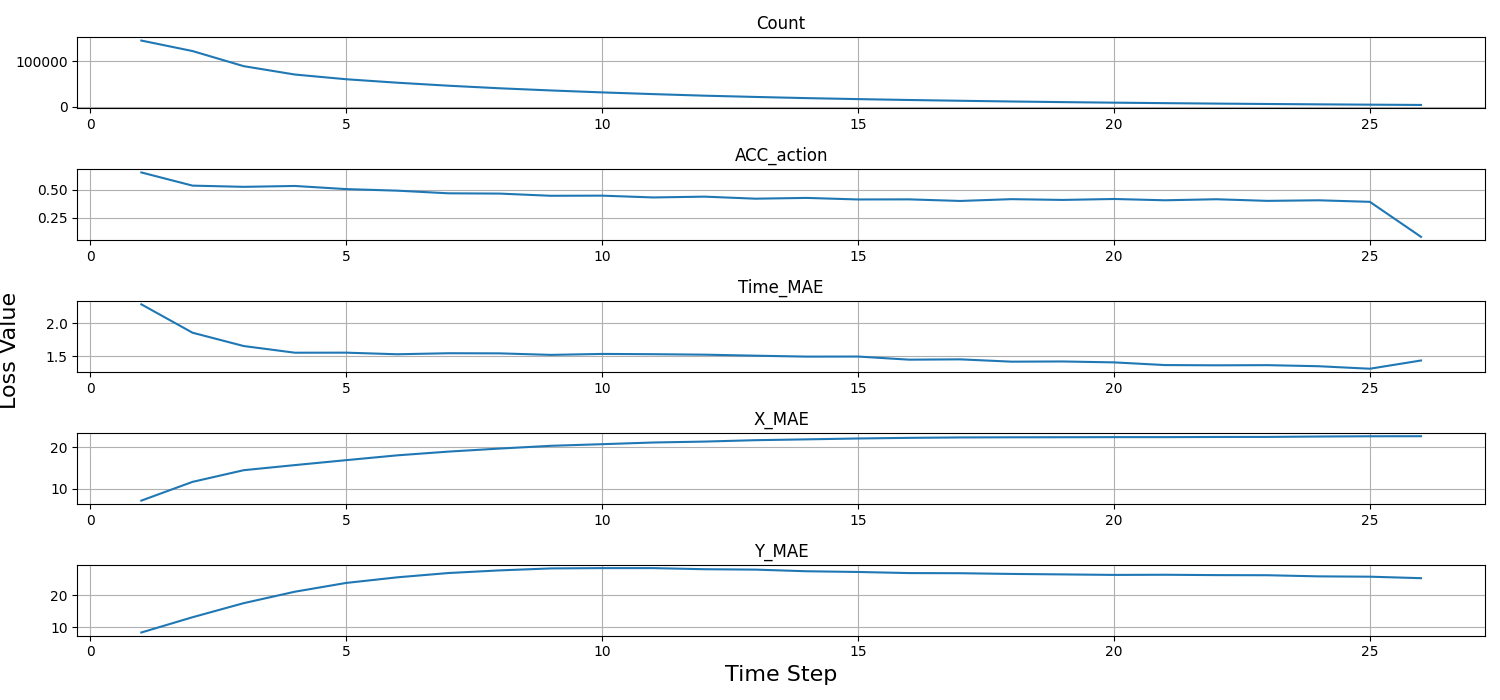}
\caption{Simulation performance of the LEM\_3 model. The exact values for the simulation are available at \url{https://github.com/open-starlab/Event/blob/main/event/sports/soccer/examples/simulation_loss.csv}}
\label{fig:simulation_eval}
\end{figure}

In this section, the simulation performance of the best-performing model LEM\_3 \citep{mendes2024forecasting} on the Statsbomb dataset (see Section \ref{ssec:dataset}) was evaluated across various metrics at each time step. The simulation utilized the greedy selection method (see Section \ref{ssec:pre-processing}), and a maximum time step of 26 was chosen, as approximately 98 percent of the possessions' number of events fall below this threshold. For each possession, if the model simulated more events than the actual number, the excess events were discarded. The evaluated metrics included event count, action classification accuracy (ACC\_action), and MAE for time and XY-coordinate predictions. The results are shown in Figure \ref{fig:simulation_eval}.

The \textbf{Event Count} decreased steadily over the time steps, reflecting a natural decay in the number of events during a soccer possession. In contrast, the \textbf{ACC\_action} dropped from 0.66 to 0.5 and remained relatively stable across most time steps, indicating that the model could consistently maintain its ability to classify 9 types of actions reasonably over time.

For the \textbf{Time\_MAE}, a noticeable decrease occurred in the early time steps, with the error stabilizing as the simulation progressed. However, this did not imply that the model improved its temporal predictions as the sequence moved forward, as the number of events decayed over time. In contrast, both the \textbf{X and Y MAEs} showed an increase over time, signaling a gradual decline in spatial prediction accuracy. The increase was more pronounced in the X MAE, suggesting that the model faced greater challenges in accurately predicting the lateral positions of events.

In summary, the simulation results demonstrated that the event model effectively maintained stable performance up until timestep 10, after which its performance worsened compared to the baseline MAJ model, as shown in Table \ref{tab:comparison_of_model}. This highlighted that while the model could predict events with reasonable accuracy in the short term, long-term predictions still had room for improvement, particularly in terms of enhancing spatial predictions.

\subsection{Dataset in RLearn Modeling}
\label{ssec:rl_dataset}
In the RL modeling experiments, J-League $2019$-$2020$ tracking and event data were used. The event data consists of a dataset that captures actions performed by players controlling the soccer ball, similar to the dataset used in Event Modeling. It records various types of actions, such as passes, shots, and fouls. For each recorded action, details are provided, including the position on the soccer pitch (given as ($x$, $y$) coordinates), event time, and action outcomes. The tracking data captures the position (represented by ($x$, $y$) coordinates) and velocity (denoted by ($v_x$, $v_y$) coordinates) of all players at all times on the soccer pitch, along with their jersey numbers and home/away status. 
This dataset contains both event and tracking data from $55$ matches of the $2019$ season Meiji J1 League soccer matches, purchased from Data Stadium Inc (see also \cite{nakahara2023action}). 

\subsection{Overview of the RLearn Modeling}
\label{ssec:rl_details}
This section explained about the overview of Rlearn modeling.
The dataset was split into training, validation, and test sets with a $50/5/45$ ratio. The data were then converted to the SAR format using the OpenSTARLab Pre-processing package (see Section \ref{ssec:pre-processing}). Afterward, all models were trained with the OpenSTARLab SAR package (see Section \ref{ssec: sar}).
The GRU, implemented using the torch library's GRU module, was employed in the reinforcement learning model following the framework of a previous study \citep{nakahara2023action}, while the MLP consists of two fully connected layers and the LSTM was implemented using the torch library.

The RL models consists of state $s$, action $a$, and reward $r$. The state $s$ consists of the position and velocity of all players and the ball, following a previous study \citep{nakahara2023action}. The action $a$ consists of a total of $16$ types: pass, through pass, shot, cross, dribble, defensive action, movement in $8$ directions (in $45$-degree increments), and idle. Since this study focuses on the evaluation of offensive players, defensive actions such as interception, tackle, clearance, and block were grouped together as a single ``defensive action'' category. Regarding the reward $r$, we followed the approach of the previous study \cite{nakahara2023action}. The reward was set to $0$ at all time steps except for the final moment of the attacking sequence. At the final time step, the reward was set to $1$ if the attacking sequence ended with a goal, $-1$ if the opponent's attacking sequence immediately afterward resulted in a conceded goal, and the expected possession value (EPV) \footnote{Calculation of EPV: \url{https://github.com/Friends-of-Tracking-Data-FoTD/LaurieOnTracking}}

In the RL models, the value function $Q(s, a)$ is updated using five elements: the current state $s$, the action $a$ selected in that state, the reward $r$ obtained for that action, the next state $s'$, and the next action $a'$ selected in that state. The discount factor $\gamma$ was set to $1.0$ in this study. This update equation is given by Equation \ref{eq:qvalue}.

\begin{align}
\label{eq:qvalue}
Q^*(s_t, a_t) = {}& E_{s_{t+1},r_{t+1}}[r_{t+1} + \gamma Q(s_{t+1},a_{t+1})|s_t,a_t] \\ \notag
= {}& \sum_{r_{t+1}} P(r_{t+1}|s_t,a_t)r_{t+1} + \gamma \sum_{s_{t+1}} P(s_{t+1}|s_t,a_t)Q(s_{t+1},a_{t+1}).
\end{align}

The loss function based on this update equation is the TD loss, expressed as follows:

\begin{align}
\label{eq:Ltd}
L_{td} = \sum^{}_{t \in T}(r_{t+1} + \gamma Q(s_{t+1}, a_{t+1}) - Q(s_{t}, a_{t}))^2.
\end{align}

Furthermore, following  \cite{nakahara2023action}, we trained the model by minimizing three loss functions: the TD loss ($L_{td}$), the action-supervised loss ($L_{as}$), and the L1 regularization loss ($L_{L1}$), as defined in Equation \ref{eq:total_loss}. The purpose of $L_{td}$ is to minimize the difference between the Q-value obtained from the current state-action pair and the actual reward obtained from the next state. $L_{as}$ aims to maximize the Q-value for the action observed in the data. $L_{L1}$ is used to prevent overfitting due to the relatively small real-world dataset. The overall loss function is defined as follows:

\begin{align}
L_{total} = L_{td} + \lambda_1L_{L1} + \lambda_2L_{as}.
\label{eq:total_loss}
\end{align}

Since $L_{td}$ aims to learn the ``optimal actions'' that maximize rewards, its optimization direction may not always align with the actual actions taken by players. On the other hand, $L_{as}$ encourages the agent to maintain specific actions. This difference creates a trade-off, making it crucial to balance the priority of these losses during the model's training process. Therefore, in our experiments, we tested different hyperparameters, $\lambda_1$ at $0.01$, $0.005$, and $0.001$, while $\lambda_2$ was set to $0.0001$, following  \citep{nakahara2023action}. Lastly, we used the Adam optimizer with a learning rate of $0.001$.

\subsection{Reinforcement learning modeling performance}
\label{ssec:rl_performance}
This section presents a comparison between the performance of the MLP, GRU, and LSTM models. Table \ref{tab:metrics_sar_comparison} summarized their action accuracy and TD loss results with different hyperparameter values. The experimental results indicate that, in a comparison of MLP and GRU, the MLP model outperformed in both action accuracy and TD loss under $\lambda_1=0.01$ and $0.005$. This suggests that while GRU has the ability to capture long-term dependencies in time-series data, simpler models like MLP might have been more suitable for tasks such as soccer.

At $\lambda_1=0.001$, all models showed similar performance, with accuracy close to the expected accuracy of $0.0625$ (Random selection of $16$ actions). This suggests that when $\lambda_1$ is small, action loss is under-optimized, causing the model to prioritize reward stability over action accuracy. This highlights the need to balance TD loss and action accuracy.
At $\lambda_1=0.005$, LSTM performed similarly to MLP and GRU at $\lambda_1=0.001$. However, LSTM’s higher complexity may lead to vanishing gradients and irrelevant information accumulation, while MLP and GRU showed better action accuracy improvements with suitable $\lambda_1$ adjustments. 

The analysis of the impact of the $\lambda_1$ revealed a trade-off relationship: increasing $\lambda_1$ improved action accuracy but raised TD loss and destabilized reward prediction, while decreasing $\lambda_1$ reduced action accuracy but lowered TD loss and enhanced reward stability. These results indicate that higher action accuracy does not necessarily improve reward prediction stability. This trade-off may stem from the reward function's incomplete representation of player behavior. 

\begin{table}[h]
\centering
\begin{tabular}{lcccc}
\toprule
Model & $\lambda_1$ & Action Acc. $\uparrow$ & TD Loss $\downarrow$  \\ \hline
\midrule
\multirow{3}{*}{MLP} & 0.01 & 0.4939 & 1.8979 \\
                    & 0.005 & 0.1674 & 0.2406 \\
                    & 0.001 & 0.0714 & 0.0001 \\ \hline
\multirow{3}{*}{GRU} & 0.01 & 0.2607 & 2.1896 \\
                    & 0.005 & 0.1311 & 0.2552 \\
                    & 0.001 & 0.0714 & 0.0001 \\ \hline
\multirow{3}{*}{LSTM} & 0.01 & 0.1982 & 0.8730 \\
                    & 0.005 & 0.0714 & 0.0001 \\
                    & 0.001 & 0.0714 & 0.0001 \\
\bottomrule
\end{tabular}
\caption{Comparison of Metrics for DataStadium. The value of $\lambda_1$ represents the hyperparameter for each model. Arrows indicate whether a higher ($\uparrow$) or lower ($\downarrow$) value is preferred for each metric.}
\label{tab:metrics_sar_comparison}
\end{table}

\subsection{Application on soccer analytics}
\label{ssec:application_on_soccer_analytics}

\begin{figure}[h!]
\centering
\begin{minipage}{0.45\textwidth}
    \centering
    \includegraphics[width=\textwidth]{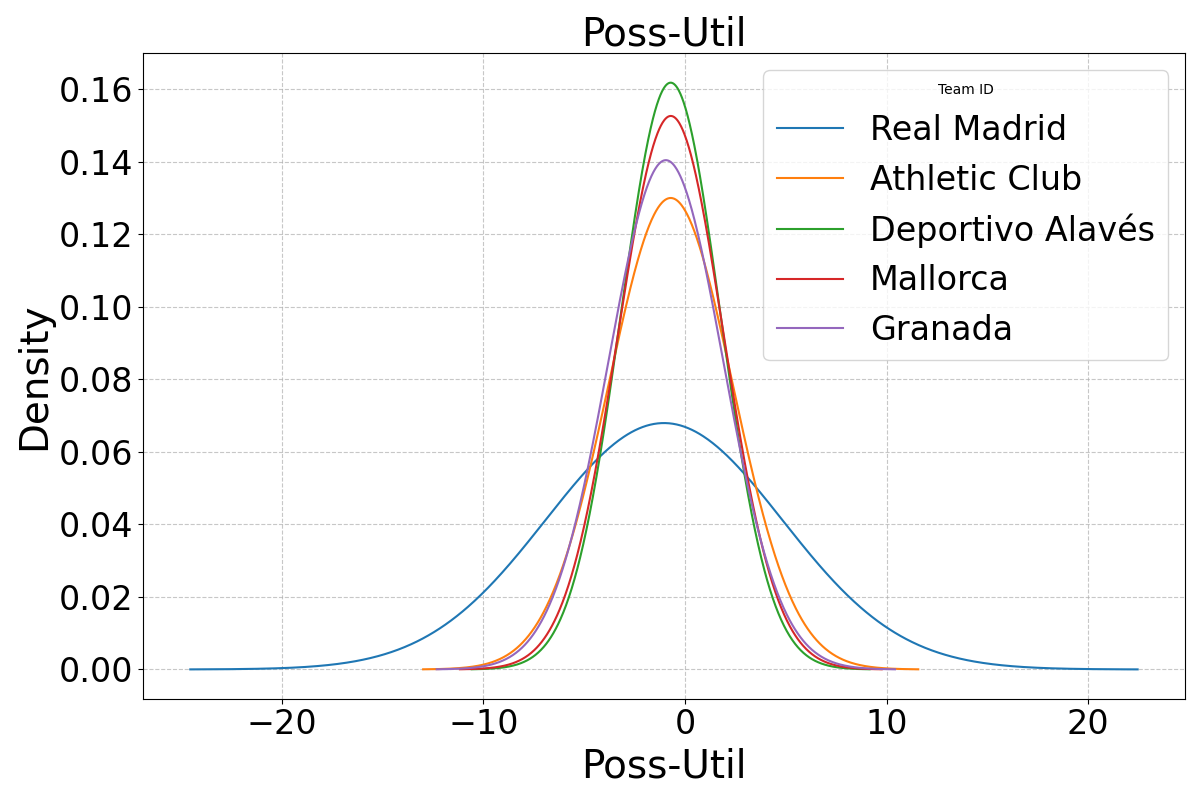}
\end{minipage}%
\hfill
\begin{minipage}{0.45\textwidth}
    \centering
    \includegraphics[width=\textwidth]{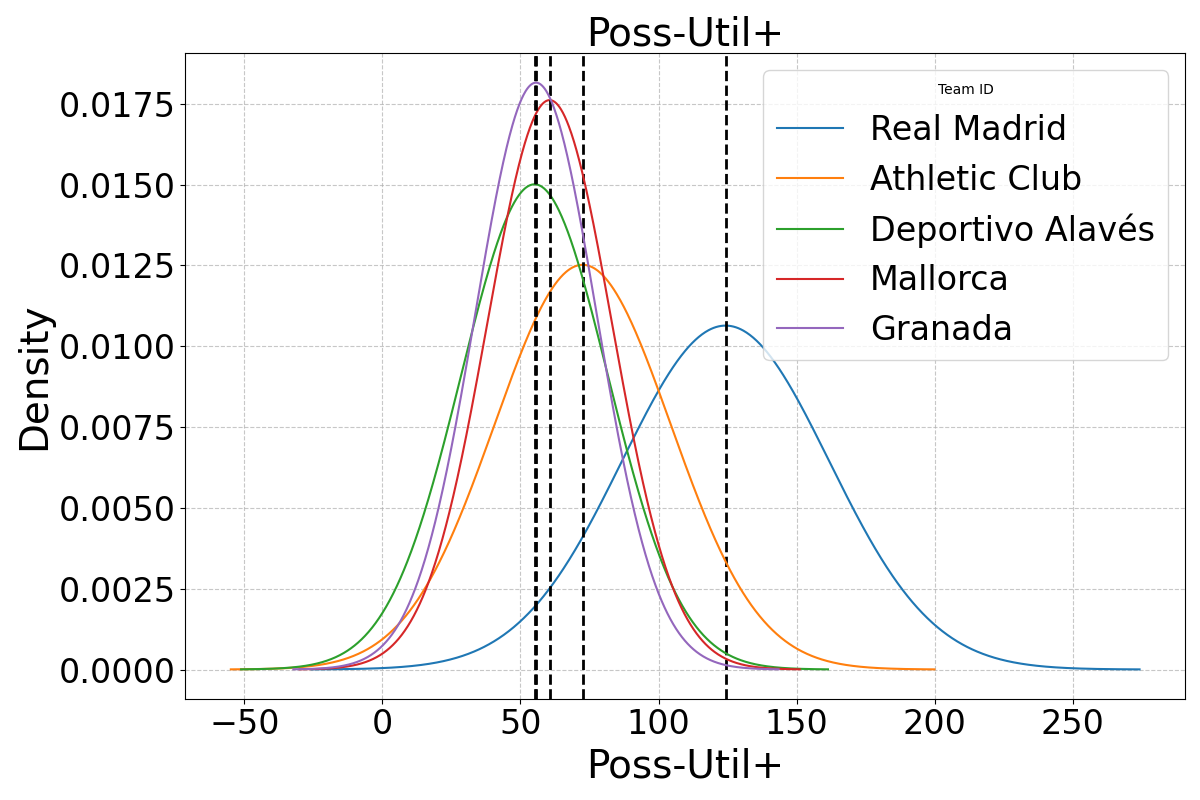} 
\end{minipage}
\caption{Poss-Util and Poss-Util+ distributions for Laliga teams. Real Madrid, Athletic Club, Deportivo Alavés, Mallorca, and Granada, ranked 1st, 5th, 10th, 15th, and 20th, respectively.}
\label{fig:poss-util}
\end{figure}

\begin{figure}[h!]
\centering
\begin{minipage}{0.5\textwidth}
    \centering
    \includegraphics[width=\textwidth]{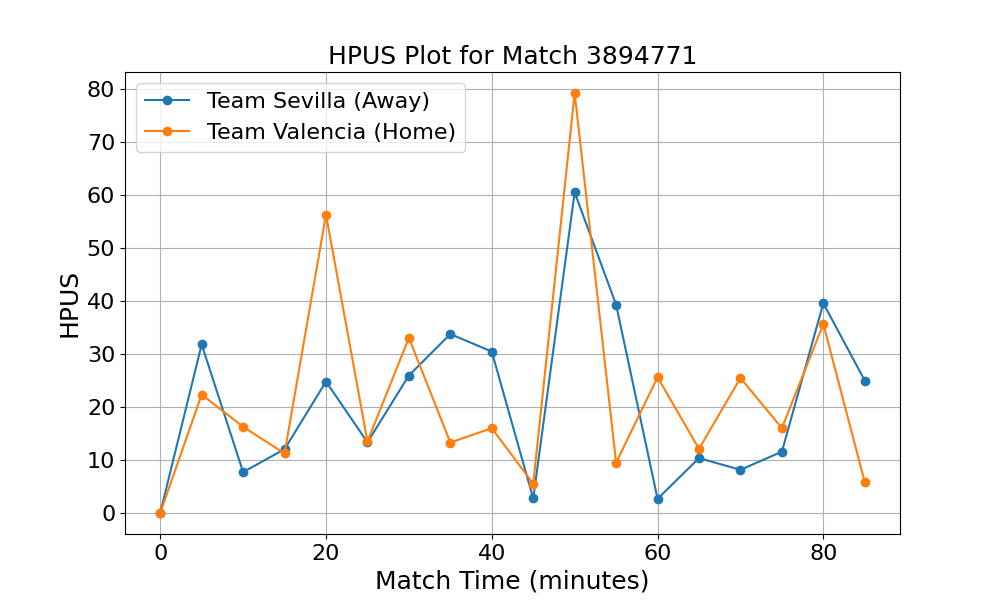}
\end{minipage}%
\hfill
\begin{minipage}{0.5\textwidth}
    \centering
    \includegraphics[width=\textwidth]{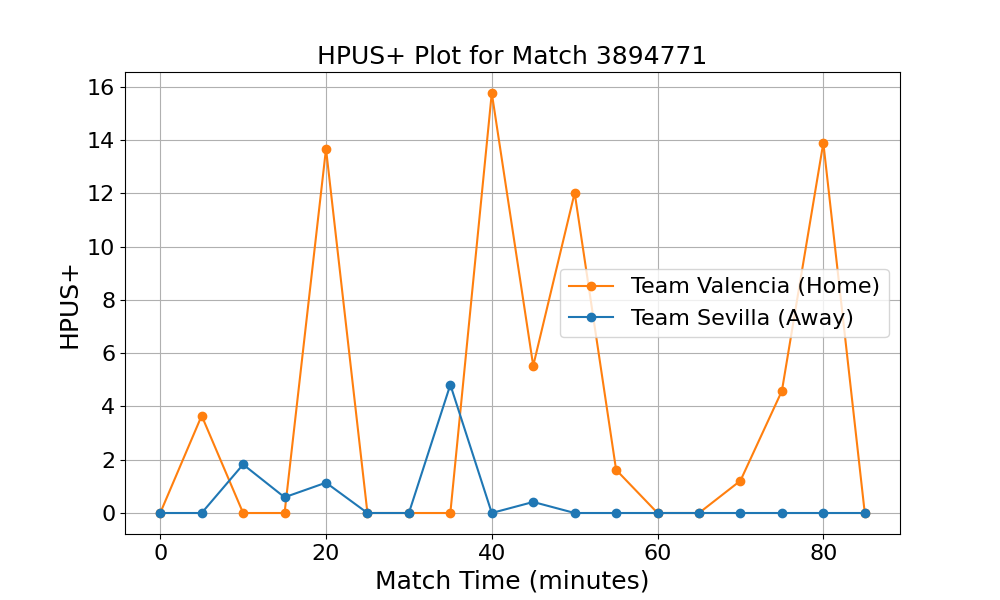}
\end{minipage}
\caption{HPUS and HPUS+ for Valencia vs. Sevilla (0-0).}
\label{fig:hpus}
\end{figure}

\begin{figure}
\centering
\rotatebox{180}{\includegraphics[width=0.5\textwidth]{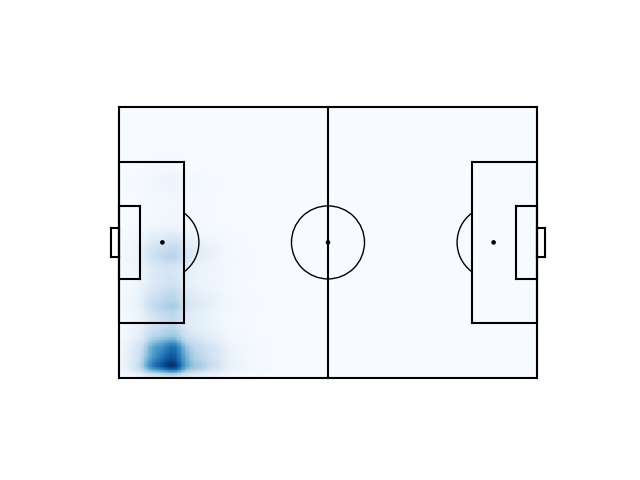}}
\caption{Heatmap of predicted cross event.}
\label{fig:heatmap}
\end{figure}

This section demonstrated the application of the \textbf{LEM\_3 model} in soccer analytics, using the test set from the StatsBomb dataset. The analysis included \textbf{Poss-Util}, \textbf{HPUS}, a heatmap of predicted events (see Section~\ref{ssec:event} for more details on the metrics), and RLearn decision-making analysis.  

In the \textbf{Poss-Util analysis} (Figure~\ref{fig:poss-util}), the Poss-Util and Poss-Util+ values were depicted for Real Madrid, Athletic Club, Deportivo Alavés, Mallorca, and Granada—teams ranked 1st, 5th, 10th, 15th, and 20th at the end of the season, respectively. In the left plot, Real Madrid demonstrated significantly higher variation in Poss-Util compared to other teams, reflecting their ability to consistently create possessions with high attacking potential. In the right plot, which focused on Poss-Util+ (positive Poss-Util values), Real Madrid displayed a higher mean and greater variation, emphasizing their superior attacking strength. In contrast, other teams showed relatively similar distributions for both Poss-Util and Poss-Util+, underscoring the gap between Real Madrid and even the 5th-ranked team.  

In the \textbf{HPUS analysis} (Figure~\ref{fig:hpus}), the match between Valencia and Sevilla, which ended in a 0-0 draw, was examined. The left plot compared the HPUS values of the two teams, showing that both had similar capabilities in utilizing possession to generate potential attacks. However, the right plot revealed that Valencia was significantly better at converting potential attacks into actual attacks than Sevilla. This highlighted Valencia’s superior attacking performance despite the match ending in a draw. Such insights provided analysts and fans with a deeper understanding of critical moments in the match and the overall performance of both teams.  

Finally, the \textbf{heatmap analysis} (Figure~\ref{fig:heatmap}) presented a visualization of predicted cross events. The heatmap helped coaches and analysts identify the spatial distribution of likely events across the field. Unlike traditional methods that relied solely on historical event data, this approach offered a broader perspective on event probabilities, enabling more informed strategic planning. By understanding the patterns of potential events, coaches could prepare more comprehensive tactics beyond specific past instances.

\begin{figure}
\centering
\includegraphics[width=1.0\textwidth]{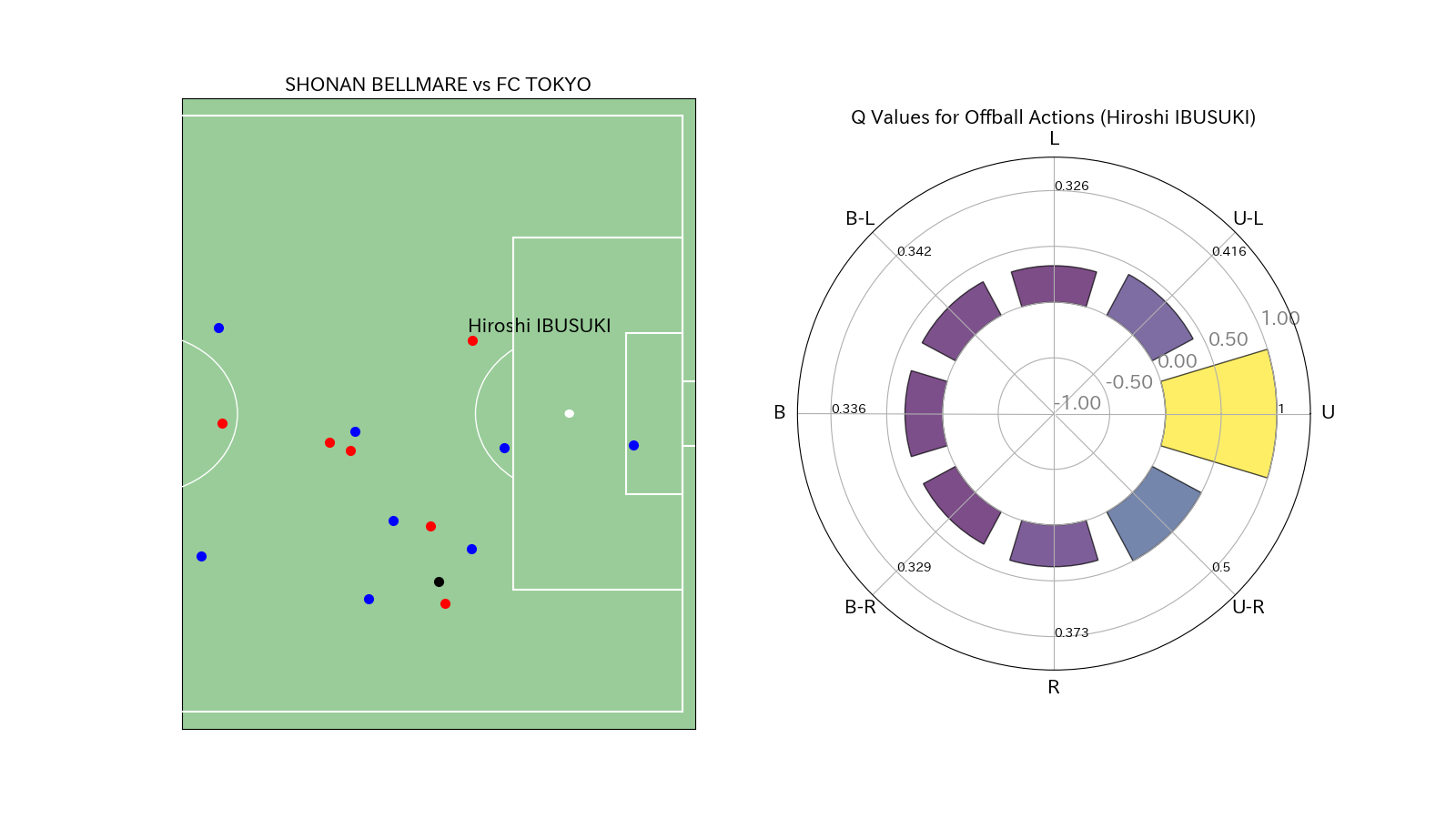}
\caption{A visualization of real-world soccer scenes and Q-values. The left figure plots the players and the ball. Attackers are plotted in red and defenders in blue. The right figure presents the Q-values (in 8 directions) of the off-ball actions of the player whose name is displayed in the left figure.}
\label{fig:RLearn_fig}
\end{figure}

Finally, the \textbf{RLearn Decision-Making Analysis} (Figure~\ref{fig:RLearn_fig}) provides a visualization of player positions and the Q-values for off-ball actions during the J1 League match between SHONAN BELLMARE and FC TOKYO. The Q-values for the players, whose names are displayed in the left diagram, are shown in the right figure. This figure is visualized in the model with $\lambda_1 = 0.05$ for the RNN, which achieved a balanced trade-off between Action Accuracy and TD error based on the experimental results \ref{tab:metrics_sar_comparison}. In this scene, forward IBUSUKI has open space in the forward direction near the goal. Given that the Q-value in front of him is also high, it can be inferred that the model appropriately predicts rewards in this situation. This visualization aids coaches and analysts in evaluating the sequential behavior of all players throughout the match.

\section{Conclusion}
\label{sec:conclusion}
In this study, we introduced OpenSTARLab, an open-source framework designed to democratize soccer spatio-temporal agent data analysis by addressing key challenges in data accessibility,  preprocessing, and modeling. This provides a Preprocessing Package including UIED (event) format from different providers and SAR format specifically tailored for DRL tasks. For cases where annotated data is unavailable, OpenSTARLab also offers an STE label Tool for annotations. In the experiments, the comparative analysis revealed that the event modeling called LEM models \citep{mendes2024towards, mendes2024forecasting} consistently outperformed other event prediction models in terms of action and time prediction accuracies. Event simulation results demonstrated that the LEM\_3 model maintained robust event prediction performance up to the tenth timesteps. RL experiments highlighted a trade-off between action accuracy and temporal difference loss. 

Overall, OpenSTARLab serves as a robust platform for researchers and practitioners, enhancing innovation and collaboration in the field of soccer data analytics.
Future research can focus on enhancing libraries for player evaluation and spatial analysis, as well as extending the framework to encompass other sports. By improving usability, we aim to broaden the applicability of OpenSTARLab, making it a versatile tool for diverse sports domains. These advancements will facilitate more comprehensive performance assessments and strategic insights across various sporting contexts, providing greater innovation in sports analytics.

\backmatter

\bmhead{Acknowledgments}
This work was financially supported by JST SPRING (Grant number JPMJSP2125), JST PRESTO (Grant number JPMJPR20CA), and NEDO Intensive Support Program for Young Promising Researchers (Grant number 24021654). The author Calvin Yeung would like to take this opportunity to thank the “Interdisciplinary Frontier Next-Generation Researcher Program of the Tokai Higher Education and Research System.” 

\section*{Declarations}
\begin{itemize}
\item Funding: This work was financially supported by JST SPRING (Grant number JPMJSP2125), JST PRESTO (Grant number JPMJPR20CA), and NEDO Intensive Support Program for Young Promising Researchers (Grant number 24021654).
\item Competing interests: The authors have no competing interests to declare that are relevant to the content of this article.
\item Ethics approval: Not applicable
\item Consent to participate: Not applicable
\item Consent for publication: Not applicable
\item Availability of data and materials: The data of the research is publicly available with details in \cite{pappalardo2019public}

\item Code availability: The code for the study is available at
\url{https://github.com/open-starlab}.
\item Authors' contributions: All authors contributed to the study conception and design. Data preparation, modeling, and analysis were performed by all authors. The first draft of the manuscript was written by all authors and all authors commented on previous versions of the manuscript. All authors read and approved the final manuscript.
\end{itemize}

\begin{appendices}
\setcounter{table}{3}
\setcounter{figure}{5}

\section{Example of training models with OpenSTARLab Event package}
\label{app:example_event}
This section provides an example to train the NMSTPP model  \citep{yeung2023transformer} and set up the YAML configuration for hyperparameter optimization using Optuna \citep{optuna_2019}. The following code snippets show how to initialize and train the NMSTPP model, followed by a YAML configuration template for optimizing hyperparameters with Optuna.
\begin{tcolorbox}[colback=gray!5!white, colframe=gray!75!black, title=Python code for training the NMSTPP model]
\begin{verbatim}
from event import Event_Model

% Initialize the NMSTPP model
model = Event_Model("NMSTPP", "path/to/NMSTPP_Configuration.yaml")

% Train the model
model.train()
\end{verbatim}
\end{tcolorbox}

\begin{tcolorbox}[colback=gray!5!white, colframe=gray!75!black, title=YAML configuration for Optuna hyperparameter optimization]
\begin{verbatim}
# OpenStarLab Event Modeling, Apache-2.0 license
# UEID data and NMSTPP model

# Training parameters
train_path: /path/to/train.csv                                                           
valid_path: /path/to/valid.csv                                                           
save_path: /path/to/save                                                                 
test: False                                                                               

num_epoch: 50
print_freq: 10
early_stop_patience: 5
dataloader_num_worker: 4
device: None                                                          

# Input features
basic_features: ['action', 'delta_T', 'start_x','start_y']
other_features: ['team','home_team','success','seconds','deltaX',
                'deltaY','distance','dist2goal','angle2goal']
use_other_features: True
num_actions: 9
seq_len: 40

# Model Hyperparameters 
# (use lists to specify hyperparameters)
optuna: True
optuna_n_trials: 100

learning_rate: [0.01]
eps: [1e-16]
batch_size: [256]

action_embedding_out_len: [9]
scale_grad_by_freq: [True]
continuous_embedding_output_len: [12] 

multihead_attention: [1] 
hidden_dim: [16,256,512,1024,2048]
feature_embedding_output_len: [21] 

NN_deltaT_num_layers: [1,2,3]
NN_location_num_layers: [1,2,3]
NN_action_num_layers: [1,2,3]
\end{verbatim}
\end{tcolorbox}




\end{appendices}

\newpage
\bibliography{sn-bibliography}

\end{document}